\pdfoutput=1

\documentclass[11pt]{article}
\usepackage[final]{acl}
\usepackage{placeins}
\FloatBarrier
\usepackage{times}
\usepackage{latexsym}
\usepackage[T1]{fontenc}
\usepackage[utf8]{inputenc}
\usepackage[whole]{bxcjkjatype}
\usepackage{microtype}
\usepackage{inconsolata}
\usepackage{hyperref}
\usepackage{url}
\usepackage{amsmath}
\usepackage{amsfonts}
\usepackage{amssymb}
\usepackage{algorithm}
\usepackage{algorithmic}
\usepackage{tcolorbox}
\usepackage{graphicx}
\usepackage{booktabs}
\usepackage{caption}
\usepackage{enumitem} 
\usepackage{graphicx}  
\usepackage{colortbl}
\usepackage{xcolor}
\usepackage{tcolorbox}
\usepackage{array}
\usepackage{caption}
\usepackage{multirow}
\usepackage{makecell}
\usepackage{todonotes}
\usepackage{tabularx}
\usepackage{float}
\definecolor{CVDblue}{HTML}{4477AA}
\definecolor{CVDorange}{HTML}{EE9944}

\newcommand{\correct}[1]{\cellcolor{CVDblue!15}{\textbf{#1}}}
\newcommand{\wrong}[1]{\cellcolor{CVDorange!25}{#1}}

\title{Diagnosing Vision Language Models' Perception by Leveraging \\ Human Methods for Color Vision Deficiencies}

\author{
Kazuki Hayashi \hspace{20pt}
Shintaro Ozaki \hspace{20pt} \\
\textbf{Yusuke Sakai} \hspace{20pt}
\textbf{Hidetaka Kamigaito} \hspace{15pt}
\textbf{Taro Watanabe} \\[4pt]
Nara Institute of Science and Technology (NAIST), Japan \\[4pt]
\texttt{hayashi.kazuki.hl4@is.naist.jp}\\
\texttt{ozaki.shintaro.ou6@naist.ac.jp}\\
\texttt{\{sakai.yusuke.sr9, kamigaito.h, taro.watanabe\}@is.naist.jp}
}

\begin{document}

\maketitle

\begin{abstract}
Large-scale Vision-Language Models (LVLMs) are being deployed in real-world settings that require visual inference. As capabilities improve, applications in navigation, education, and accessibility are becoming practical. These settings require accommodation of perceptual variation rather than assuming a uniform visual experience. Color perception illustrates this requirement: it is central to visual understanding yet varies across individuals due to Color Vision Deficiencies, an aspect largely ignored in multimodal AI.
In this work, we examine whether LVLMs can account for variation in color perception using the Ishihara Test. We evaluate model behavior through generation, confidence, and internal representation, using Ishihara plates as controlled stimuli that expose perceptual differences. Although models possess factual knowledge about color vision deficiencies and can describe the test, they fail to reproduce the perceptual outcomes experienced by affected individuals and instead default to normative color perception. These results indicate that current systems lack mechanisms for representing alternative perceptual experiences, raising concerns for accessibility and inclusive deployment in multimodal settings.
\end{abstract}

\section{Introduction}
\label{introduction}

Large-scale Vision-Language Models (LVLMs) \cite{liu2024llavanext, abdin2024phi3technicalreporthighly, ye2024mplugowl3longimagesequenceunderstanding, bai2025qwen25vltechnicalreport} extend language models with visual perception and enable unified multimodal reasoning. As LVLMs move from research prototypes to real-world interactive systems, they are increasingly deployed in navigation, education, and assistive technologies \cite{pmlr-v229-zitkovich23a, jimaging10050103, morita2024contextawaresupportcolorvision, zhou2025autovla}. In such applications, LVLMs serve as a medium through which users access and interpret visual information in their environment.

\begin{figure}[t]
\centering
\includegraphics[width=1.00\columnwidth]{images/figure_1.pdf}
\caption{An example of communication failure due to CVDs: a user with CVD may not correctly interpret the model's color-based instructions. This highlights the need for LVLMs to account for perceptual differences in real-world interactions.}
\label{fig:intro}
\end{figure}

A key yet underexplored assumption in such deployments is that users share a similar visual experience of the world. In reality, visual perception varies across individuals. 
Color vision in particular shows substantial variability due to biological factors \cite{Emery2019-hi}, and Color Vision Deficiencies (CVDs) are a common form of such diversity, impairing discrimination between colors such as red and green \cite{Birch2012-xg, roberson2007color}. These differences are not only perceptual but also functional: color vision affects wayfinding, diagram and chart reading, medical interpretation, and user interface navigation. As illustrated in Figure~\ref{fig:intro}, an LVLM-based navigation assistant instructing a traveler to ``Take the green line'' may fail to provide actionable information for a red–green color blind user, raising concerns about inclusion, accessibility, and fairness in real deployments \cite{Kawakita2024}. Systems designed around a ``standard'' visual experience may silently exclude users with different perceptual characteristics, even when linguistic content remains fully accessible.

Current LVLMs are trained largely on web-scale image-text corpora that implicitly assume typical color vision \cite{paik-etal-2021-world, samin-etal-2025-colorfoil, Rahmanzadehgervi_2024_ACCV}. As a result, these models acquire substantial knowledge about color, color blindness, and clinical diagnostic tools. However, whether such knowledge enables the model to simulate alternative visual percepts remains unclear. This distinction is central: explaining what CVD is in language is not the same as perceiving colors as a CVD user would. To our knowledge, the ability of LVLMs to reproduce perceptual diversity has not been systematically investigated.

To address this, we leverage the Ishihara Test \cite{ishihara1917test}, a clinically validated tool widely used for diagnosing CVDs in humans. Rather than using Ishihara solely as a screening instrument, we repurpose it as a \textit{diagnostic probe} for LVLMs. Because Ishihara plates are designed to elicit different perceptual responses across color vision types, they allow direct comparison between model outputs and established human perceptual categories. 
In addition, the test provides a standardized visual stimulus that isolates color-based perceptual variation without cultural or linguistic interference, and its clinical validation over a century allows perceptual differences to be evaluated without new annotation pipelines or subjective rating procedures.

We evaluate LVLM behavior through three complementary perspectives. At the \textit{generation level}, we assess whether the model outputs the digit perceived by users with different CVD types. At the \textit{confidence level}, we analyze perplexity under simulated CVD conditions to determine whether uncertainty patterns align with human perceptual tendencies. At the \textit{representation level}, we apply a layer-wise LogitLens probing~\cite{nostalgebraist2020logitlens} method to assess whether internal activations encode distinctions between vision types. Together, these perspectives allow us to analyze how perceptual variation interacts with LVLMs' generative, probabilistic, and latent processing mechanisms.

Our findings show that although LVLMs possess strong linguistic knowledge about CVDs and can describe the Ishihara Test, they fail to reproduce the perceptual outcomes experienced by individuals with CVD. Even when given additional linguistic descriptions or visual exemplars, the models default to normative color perception, indicating insufficient alignment between linguistic knowledge and perceptual grounding. These results suggest that current LVLMs lack mechanisms for representing alternative perceptual experiences, which has implications for accessibility, fairness, and human-centered deployment in multimodal AI as such systems increasingly mediate visual information.

\section{Background}

\subsection{Color Vision Deficiency and Ishihara Test}
Color vision in humans relies on three types of cone photoreceptors: red, green, and blue, each sensitive to long, medium, or short wavelengths. 
CVDs affect about 8\% of males and 0.5\% of females \cite{Birch2012-xg}. When one type of cone is missing or doesn't work properly, large parts of the color spectrum lose their typical hue or brightness, resulting in CVDs. 
The three main types are summarized below:

\begin{description}
  \setlength\itemsep{2pt}
  \item[Protanopia] Absence of long wavelength (L) cones makes reds appear very dark and compresses the differences among many reds, greens, and browns; purples are often mistaken for blue.  
  \item[Deuteranopia] Absence of medium wavelength (M) cones leaves brightness almost unaffected but shifts greens toward beige, producing strong red and green confusion.  
  \item[Tritanopia] Absence of short wavelength (S) cones narrows the blue to yellow axis; blues drift toward green, cyans toward gray, and yellows toward pink or light gray, greatly reducing blue versus yellow separability.  
\end{description}

\begin{table}[t]
    \centering
    \small
    \renewcommand{\arraystretch}{1.00}  
    \begin{tabular}{ccc}
        \toprule
        \textbf{Plate} & \textbf{Type} & \textbf{Answer} \\
        \midrule
        \multirow{4}{*}{\centering\includegraphics[width=0.18\linewidth]{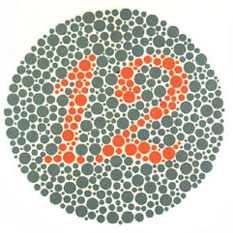}} &  Normal & 12 \\
        & Protanopia & 12 \\
        & Deuteranopia & 12 \\
        & Tritanopia & 12 \\
        \midrule
        \multirow{4}{*}{\centering\includegraphics[width=0.18\linewidth]{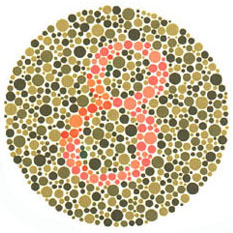}} & Normal & 8 \\
        & Protanopia & 3 \\
        & Deuteranopia & 3 \\
        & Tritanopia & 8 \\
        \midrule
        \multirow{4}{*}{\centering\includegraphics[width=0.18\linewidth]{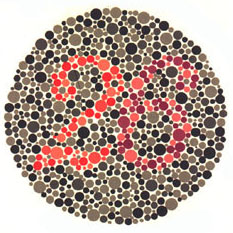}} & Normal & 26 \\
        & Protanopia & 6 \\
        & Deuteranopia & 2 \\
        & Tritanopia & 26 \\        
        \bottomrule
    \end{tabular}
    \caption{Ground-truth answers for each Ishihara plate under different vision conditions, based on clinical interpretations of the Ishihara color test~\cite{ishihara1917test}.}
    \label{tab:ishihara_examples}
\end{table}

To screen for CVDs, the Ishihara Test \citep{ishihara1917test} has been widely used for decades in clinical and research settings~\cite{Fanlo_Zarazaga2019-wc}. Each plate shows a figure made of colored dots. People with normal vision usually identify it, while those with CVDs may see a different number or nothing.  For instance, the plate in Table~\ref{tab:ishihara_examples} looks like ``8'' to a person with normal vision, but is often read as ``3'' by people with red and green deficiencies \cite{Birch2012-xg}. By using typical color confusions across CVDs, the Ishihara Test remains a fast and reliable tool for identifying these conditions.

\subsection{CVDs and Ishihara Tests in Recent Research}
Despite its limited size, the Ishihara Test remains a clinically validated and reliable tool for CVD detection.
\citet{CHEN1995463} applied automated analysis to recognize Ishihara patterns, enabling automated color vision testing.
Later work confirmed that digital versions retain reliability across devices \cite{zhao2019comparison, klinke2024comparative}. 
\citet{computers14020034, grover2025huemanityprobingfinegrainedvisual} introduced synthetic Ishihara-style datasets to assess model performance on digit or character recognition under red and green confusion.
CVDs compensation methods also use Ishihara principles to enhance accessibility for people with color vision deficiency. 
Key approaches include physiologically based daltonization~\cite{5290741}, eigenvector-based recoloring~\cite{s19102250}, and neural real-time Ishihara recoloring~\cite{inbook}.
These works highlight Ishihara plates as diagnostic tools and foundations for assistive color algorithms.

\begin{table}[t]
  \centering
  \small
  \resizebox{1.00\columnwidth}{!}{
    \begin{tabular}{cp{0.73\columnwidth}}
      \toprule
      \textbf{Input} & \textbf{Extracted mPLUG-Owl3 Output} \\
      \midrule
      \multirow{2}{*}{\shortstack{What is\\Ishihara Test?}} &
      … visual test \textbf{used to detect color-vision deficiencies, especially red–green types such as protanopia and deuteranopia.} It presents a pattern of red and green dots forming a hidden number; individuals with normal vision can read the number, whereas those with certain CVDs cannot. \\[2pt]
      \midrule
      \multirow{2}{*}{\shortstack{What is\\Protanopia?}} &
      … \textbf{color blindness that affects a person's ability to distinguish between red and green colors.} In people with protanopia, the cone cells in the retina that are responsible for perceiving red and green colors are not functioning properly. \\[2pt]
      \bottomrule
    \end{tabular}}
  \caption{Excerpts from mPLUG-Owl3 showing knowledge of the Ishihara test and CVDs. The full outputs are available in Appendix~\ref{appendix:cvd_knowledge}.}
\label{tab:mplug_cvd_descriptions}
\end{table}

\subsection{Color Understanding in LVLMs}
Early VQA models~\cite{balanced_vqa_v2} explored color-related questions but relied on priors over visual understanding, yielding inconsistent results.
\citet{Agrawal_2018_CVPR,Kervadec_2021_CVPR} proposed debiased datasets to reduce language bias and improve visual grounding.
Transformer-based models~\cite{pmlr-v139-radford21a,pmlr-v139-kim21k} enhanced image-text alignment and performed well on some tasks yet still struggle with fine-grained color distinctions~\cite{samin-etal-2025-colorfoil}.
Although recent LVLMs exhibit strong visual reasoning, their color understanding remains limited by insufficient supervision for subtle or ambiguous cases.
\citet{burapacheep2024colorswapcolorwordorder, shahgir2024illusionvqachallengingopticalillusion} found that models struggle with basic color-word grounding and are easily misled by perceptual illusions.
\citet{liang2025colorbenchvlmsunderstandcolorful} evaluated diverse color tasks and showed even large models struggle with shifts and ambiguity, revealing persistent limitations.

\section{Experimental Setup}

\subsection{Pilot Study}
\label{preliminary-experiments}
Before analyzing our main question of whether LVLMs can simulate aspects of human color perception, we first verified that they possess basic language-level knowledge about color vision deficiencies and the Ishihara Test.
To test this prerequisite, we used two types of prompts: (i) asking an Ishihara plate ``\texttt{What is this image used for?}'', and (ii) requesting definitions such as ``\texttt{What is Protanopia as a type of color vision deficiency?}''. As shown in Table~\ref{tab:mplug_cvd_descriptions}, all evaluated models were judged to respond correctly. This confirms that they satisfy the necessary prerequisite knowledge for the analyses that follow. 
Appendix~\ref{appendix:cvd_knowledge} provides full outputs.

\subsection{Research Questions}
\label{sec:experiments}
We use the Ishihara Test as with humans to analyze how LVLMs simulate color vision across task settings\footnote{We use 25 out of 38 plates that require numeral responses to ensure consistent analysis across visual conditions.}.
In our research, ``color perception'' means the model's ability to respond like humans with specific CVDs, rather than identifying colors.
We investigate this across three levels: generation, confidence, and internal representation.

\begin{table}[t]
    \centering
    \resizebox{1.00 \columnwidth}{!}{
    \small
    \renewcommand{\arraystretch}{0.80}  
    \begin{tabular}{p{0.18\columnwidth} p{0.75\columnwidth}}
        \toprule
        \textbf{Type} & \textbf{Prompt} \\
        \midrule

        \textbf{Base} &
        \textcolor{blue}{\texttt{\{Image: Ishihara plate\}}} \newline
        You are Protanopic. \newline
        What number do you see?
        Only provide the answer. No additional information. \\
        
        \midrule
        \multirow{2}{*}{\shortstack{\textbf{Linguistic} \\ \textbf{Support}}} & 
        \textcolor{blue}{\texttt{\{Image: Ishihara plate\}}} \newline
        You are Protanopic. \newline
        Protanopia means you have difficulty distinguishing red tones, as red appears dimmer and may be confused with green or brown. \newline
        What number do you see?
        Only provide the answer. No additional information. \\
        
        \midrule
        \multirow{2}{*}{\shortstack{\textbf{Visual} \\ \textbf{Support}}} & 
        \textcolor{blue}{\texttt{\{Reference Image: Ishihara plate\}}} \newline
        You are Protanopic. \newline
        Example 1:
        Image shows an Ishihara plate.
        A Protanopic person sees this number: 2. \newline
        \textcolor{blue}{\texttt{\{Image: Ishihara plate\}}} \newline
        Now, look at the following image. 
        What number do you see? 
        Only provide the answer. No additional information. \\
        
        \midrule
        \textbf{Doctor-Style} & 
        \textcolor{blue}{\texttt{\{Image: Ishihara plate\}}} \newline
        When a person sees “\{task\_number\}” in this Ishihara test plate, which type of color vision is most likely? \newline
        Choose from the following options: \newline
        - Normal
        - Protanopia
        - Deuteranopia
        - Tritanopia
        Please output your answer. \\

        \bottomrule
    \end{tabular}
    }
    \caption{
    Examples of prompts used in our experiments. 
Base, Linguistic, and Visual prompts are used in RQ1-RQ2 for digit generation, 
and Doctor-Style is used in RQ3 for vision-type inference. 
All include an Ishihara plate as visual input.
    }
    \label{tab:representative_prompts}
\end{table}

\begin{description}
\item[RQ1 (Generation level).]
Can LVLMs produce Ishihara digit responses that align with those reported by individuals with color vision deficiency types?
This evaluates whether the model’s generative outputs can emulate alternative perceptual outcomes rather than defaulting to neurotypical perception.

\item[RQ2 (Confidence level).]
Do model confidence patterns vary across vision conditions, and do uncertainty signals (measured via perplexity) reflect condition-specific perceptual ambiguity observed in humans?
This probes whether LVLMs assign different likelihoods to Ishihara responses depending on simulated vision type.

\item[RQ3 (Representation level).]
Do internal layerwise activation patterns encode distinctions consistent with color vision deficiency categories?
This examines whether alternative perceptual states are represented at the latent level, even if not expressed in the final output.
\end{description}

\subsection{Task Definition}
The task is inspired by the human Ishihara Test: given an Ishihara plate, the model is instructed to simulate a target vision condition and report the perceived digit. We simulate four conditions: \{Normal, Protanopia, Deuteranopia, Tritanopia\}. Since the Ishihara Test does not assess Tritanopia, the correct digits for Normal and Tritanopia coincide. We exploit this property by treating Tritanopia as a \emph{control condition}: if a model meaningfully incorporates the instruction ``\texttt{You are Tritanopic},'' its output distribution or confidence should differ from Normal despite identical ground-truth digits. As shown later, most LVLMs instead return nearly identical responses for the two conditions, suggesting weak sensitivity to instructed perceptual states.

\paragraph{Prompts}
\label{appendix:prompts}
To analyze whether LVLMs can simulate CVDs under different instructional prompts, we use Ishihara plates and design four types of prompts: Base, Linguistic Support, Visual Support and Doctor-Style, as shown in Table~\ref{tab:representative_prompts}.
We tested multiple variants within each prompt type but observed no substantial differences. Thus, we use one representative set in our main experiments and report the variant analysis in Appendix~\ref{appendix:prompt_variants}.

\begin{description}
\item[Base Prompts] 
These provide only the condition, e.g., ``You are Protanopic'', and ask the model to report the number in the Ishihara plate. This tests whether the model can simulate CVD perception specified by the condition.

\item[Linguistic Support]  
In addition to the Base prompt, these add a short description of the impairment, e.g.,  ``red tones appear darker,'' to assess whether textual context aids prediction.

\item[Visual Support]
These include a brief condition description and a reference example showing what number appears in another plate, before asking the model to identify the number in a new image. This tests whether few-shot visual examples help simulate CVD perception.

\item[Doctor-Style]  
Instead of generating digits, the model infers the vision type (Normal, Protanopia, Deuteranopia, or Tritanopia) from a given response.  
Used in RQ3, this prompt avoids tokenization issues with multi-digit numerals (e.g., ``18'' $\rightarrow$ ``1'', ``8'') by using diagnosis labels, which are consistently tokenized and clearly distinguishable across models.

\end{description}

\begin{table*}[t]
  \centering
  \resizebox{1.00\textwidth}{!}{%
    \begin{tabular}{l r  rrr | rrr | rrr | rrr | rrr}
      \toprule
       & 
        & \multicolumn{3}{c}{\textbf{Normal}}
        & \multicolumn{3}{c}{\textbf{Protanopia}}
        & \multicolumn{3}{c}{\textbf{Deuteranopia}}
        & \multicolumn{3}{c}{\textbf{Tritanopia}}
        & \multicolumn{3}{c}{\textbf{Avg.}} \\
      \cmidrule{3-5} \cmidrule{6-8} \cmidrule{9-11} \cmidrule{12-14} \cmidrule{15-17}
        \multirow{-2}{*}{\textbf{LVLM}} &  \multirow{-2}{*}{\textbf{Size}}
        & \textbf{Base } & \textbf{Ling.} & \textbf{Vis. }
        & \textbf{Base } & \textbf{Ling. } & \textbf{Vis. }
        & \textbf{Base } & \textbf{Ling. } & \textbf{Vis. }
        & \textbf{Base } & \textbf{Ling. } & \textbf{Vis. }
        & \textbf{Base } & \textbf{Ling.} & \textbf{Vis. } \\
      \midrule
      Llama-3.2   & 11B  & 14.3  & 9.5   & --  & 5.9   & 5.9   & --  & 11.8  & 5.9   & --  & 4.8   & 23.8  & --  & 9.2   & 11.3  & -- \\
      LLaVA-NeXT   & 13B  & 52.4  & 52.4  & --  & 5.9   & 5.9   & --  & 5.9   & 5.9   & --  & 52.4  & 61.9  & --  & 29.2  & 31.5  & -- \\
      mPLUG-Owl3   & 7B   & 71.4  & 71.4  & 62.0 & 5.9   & 5.9   & 5.9 & 5.9   & 5.9   & 17.7 & 66.7  & 66.7  & 62.0 & 37.5  & 37.5  & 36.9 \\
      Phi-3.5     & 4.2B & 0.0   & 0.0   & 4.8  & 0.0   & 0.0   & 5.9 & 0.0   & 0.0   & 11.8 & 0.0   & 0.0   & 9.5  & 0.0   & 0.0   & 8.0 \\
      Qwen2.5-VL     & 7B   & 28.6  & 0.0   & 4.8  & 0.0   & 0.0   & 5.9 & 0.0   & 0.0   & 0.0  & 23.8  & 0.0   & 4.8  & 13.1  & 0.0   & 3.9 \\
      \midrule
      GPT-4o  & --   & 90.5  & 90.5  & 90.5 & 23.6  & 17.7  & 23.6& 5.9   & 23.5  & 5.9  & 71.4  & 66.7  & 42.9 & 47.9  & 48.4  & 40.7 \\
      \bottomrule
    \end{tabular}%
  }
\caption{\emph{Digit Accuracy} (\%) of each LVLM under simulated Vision types: \textbf{Normal} and three CVD types, with support settings: \textbf{Base}, \textbf{Ling.} (Linguistic), and \textbf{Vis.} (Visual). 
\textbf{Vis.} was used only for models supporting multi-image input. 
Plates for which no correct digit exists (marked as ``N/A''in the Gold answers in Table~\ref{tab:output_digit_base_all}) were excluded from the accuracy calculation.}
\label{tab:result-score}
\end{table*}

\subsection{Evaluation Metrics}
\label{sec:metrics}

We quantify each research question using a dedicated and interpretable metric for each aspect.

\paragraph{Digit Accuracy (RQ1)}
The proportion of Ishihara plates on which the model's generated digit
matches the ground‑truth numeral under the specified vision condition.

\paragraph{Per‑Token Perplexity (RQ2)}
For each plate, we concatenate the image, prompt, and gold numeral
answer into a single input sequence.  
All tokens preceding the answer are masked with
\texttt{ignore\_index}, ensuring that only the answer tokens contribute
to the loss.  
The answer consists of \(L\) tokens, denoted as \(\boldsymbol{x}_{1:L}\).  We compute their token‑level
log‑probabilities
\[
v_i = \log P\left(x_i \mid \textit{prompt},\,\boldsymbol{x}_{1:i-1}\right),
\]
and the average negative log‑likelihood
\[
\mathcal{L} \;=\; -\frac{1}{L}\sum_{i=1}^{L} v_i,
\quad
\mathrm{PPL} \;=\; \exp(\mathcal{L}).
\]
A lower \(\mathrm{PPL}\) indicates higher confidence in the forced-decoded answer.

\paragraph{Layer‑Wise Diagnosis Probability (RQ3)}
This approach builds on prior work showing that intermediate representations influence model predictions~\cite{wendler-etal-2024-llamas, schut2025multilingual}.
At each Transformer layer, we project the hidden state \(\boldsymbol{h}\) into
the vocabulary space using the unembedding matrix \(\mathbf{W}^{\text{unembed}}\)
and apply a softmax:
\[
\mathrm{LogitLens}(\boldsymbol{h}) = \mathrm{softmax}\left(\mathbf{W}^{\text{unembed}}\,\boldsymbol{h}\right).
\]
The resulting distribution gives the probability of every token at
that layer~\cite{nostalgebraist2020logitlens}.  For a specific diagnosis‑label token \(t\)
(e.g., \textit{Protanopia} or \textit{Deuteranopia}) we record
\[
P(t) =
\frac{\exp\left(\boldsymbol{W}_t^{\text{unembed}} \cdot \boldsymbol{h}\right)}
     {\sum_{j} \exp\left(\boldsymbol{W}_j^{\text{unembed}} \cdot \boldsymbol{h}\right)}.
\]
Plotting \(P(t)\) across layers reveals how strongly the model favors
each color‑vision diagnosis while processing the plate, offering insights into its internal decision-making process.

\subsection{Models}
We evaluate six LVLMs:
Phi-3.5~\cite{abdin2024phi3technicalreporthighly},
Qwen2.5-VL~\cite{bai2025qwen25vltechnicalreport},
mPLUG-Owl3~\cite{ye2024mplugowl3longimagesequenceunderstanding},
LLaVA-NeXT~\cite{liu2024llavanext},
Llama-3.2~\cite{grattafiori2024llama3herdmodels},
and GPT-4o~\cite{openai2024gpt4ocard}.
GPT-4o is proprietary; the remaining five are open-source and were selected for passing the Pilot Study prerequisite.  
Appendix~\ref{appendix:Detailed Settings} provides the detailed settings.

\begin{table*}[ht]
\centering
\small
\resizebox{1.00\textwidth}{!}{%
  \renewcommand{\arraystretch}{1.2}
  \begin{tabular}{l*{26}{c}}
    \toprule
    \multirow{7}{*}{Normal}
      & \cellcolor{blue!10}\textbf{Gold}
      & \textbf{12}&\textbf{8}&\textbf{6}&\textbf{29}&\textbf{57}&\textbf{5}&\textbf{3}&\textbf{15}&\textbf{74}
      & \textbf{2}&\textbf{6}&\textbf{97}&\textbf{45}&\textbf{5}&\textbf{7}&\textbf{16}&\textbf{73}
      & \textbf{N/A}&\textbf{N/A}&\textbf{N/A}&\textbf{N/A}&\textbf{26}&\textbf{42}&\textbf{35}&\textbf{96}\\
    \cmidrule(lr){2-27}
      & Llama
      & \wrong{19}&\correct{8}&\wrong{8}&\wrong{42}&\wrong{7}&\wrong{1}&\wrong{2}&\wrong{4}&\wrong{15}
      & \correct{2}&\wrong{42}&\wrong{42}&\wrong{1}&\wrong{7}&\correct{7}&\wrong{4}&\wrong{53}
      & \wrong{15}&\wrong{15}&\wrong{3}&\wrong{4}&\wrong{52}&\wrong{52}&\wrong{52}&\wrong{8}\\
      & LLaVA
      & \correct{12}&\correct{8}&\correct{6}&\correct{29}&\wrong{37}&\wrong{3}&\wrong{6}&\wrong{16}&\wrong{7}
      & \correct{2}&\correct{6}&\wrong{9}&\wrong{5}&\wrong{6}&\wrong{7}&\correct{16}&\wrong{3}
      & \wrong{1}&\wrong{1}&\wrong{1}&\wrong{1}&\wrong{28}&\correct{42}&\correct{35}&\correct{96}\\
      & mPLUG
      & \correct{12}&\correct{8}&\correct{6}&\correct{29}&\wrong{37}&\correct{5}&\correct{3}&\correct{15}&\wrong{24}
      & \correct{2}&\correct{6}&\wrong{9}&\wrong{10}&\correct{5}&\correct{7}&\correct{16}&\wrong{13}
      & \wrong{100}&\wrong{100}&\wrong{10}&\wrong{100}&\wrong{23}&\correct{42}&\correct{35}&\correct{96}\\
      & Phi
      & --&--&--&--&--&--&--&--&--
      & --&--&--&--&--&--&--&--
      & --&--&--&--&--&--&--&--\\
      & Qwen
      & \correct{12}&\correct{8}&\correct{6}&\correct{29}&\correct{57}&\wrong{74}&\wrong{74}&\wrong{74}&\correct{74}
      & \wrong{74}&\wrong{74}&\wrong{74}&\wrong{74}&\wrong{74}&\wrong{74}&\wrong{74}&\wrong{74}
      & \wrong{74}&\wrong{74}&\wrong{74}&\wrong{74}&\wrong{74}&\wrong{48}&\wrong{74}&\wrong{95}\\
      & GPT
      & \correct{12}&\correct{8}&\correct{6}&\correct{29}&\correct{57}&\correct{5}&\correct{3}&\correct{15}&\correct{74}
      & \correct{2}&\correct{6}&\wrong{74}&\correct{45}&\correct{5}&\correct{7}&\correct{16}&\correct{73}
      & \wrong{74}&\wrong{74}&\wrong{74}&\wrong{74}&\wrong{5}&\correct{42}&\correct{35}&\correct{96}\\
    \midrule
    \multirow{7}{*}{Protanopia}
      & \cellcolor{blue!10}\textbf{Gold}
      & \textbf{12}&\textbf{3}&\textbf{5}&\textbf{70}&\textbf{35}&\textbf{2}&\textbf{5}&\textbf{17}&\textbf{21}
      & \textbf{N/A}&\textbf{N/A}&\textbf{N/A}&\textbf{N/A}&\textbf{N/A}&\textbf{N/A}&\textbf{N/A}&\textbf{N/A}
      & \textbf{5}&\textbf{2}&\textbf{45}&\textbf{73}&\textbf{6}&\textbf{2}&\textbf{5}&\textbf{6}\\
    \cmidrule(lr){2-27}
      & Llama
      & \wrong{19}&\wrong{2}&\correct{5}&\wrong{17}&\wrong{3}&\wrong{3}&\wrong{2}&\wrong{6}&\wrong{3}
      & \wrong{2}&\wrong{2}&\wrong{2}&\wrong{45}&\wrong{7}&\wrong{4}&\wrong{6}&\wrong{46}
      & \wrong{73}&\wrong{4}&\wrong{53}&\wrong{52}&\wrong{4}&\wrong{4}&\wrong{4}&\wrong{2}\\
      & LLaVA
      & \correct{12}&\wrong{8}&\wrong{6}&\wrong{29}&\wrong{57}&\wrong{3}&\wrong{6}&\wrong{16}&\wrong{7}
      & \wrong{2}&\wrong{6}&\wrong{9}&\wrong{1}&\wrong{6}&\wrong{7}&\wrong{16}&\wrong{3}
      & \wrong{1}&\wrong{1}&\wrong{1}&\wrong{1}&\wrong{20}&\wrong{42}&\wrong{38}&\wrong{96}\\
      & mPLUG
      & \correct{12}&\wrong{8}&\wrong{6}&\wrong{29}&\wrong{37}&\wrong{5}&\wrong{3}&\wrong{15}&\wrong{24}
      & \wrong{9}&\wrong{6}&\wrong{9}&\wrong{10}&\wrong{5}&\wrong{7}&\wrong{16}&\wrong{13}
      & \wrong{1}&\wrong{1}&\wrong{1}&\wrong{1}&\wrong{23}&\wrong{42}&\wrong{35}&\wrong{96}\\
      & Phi
      & --&--&--&--&--&--&--&--&--
      & --&--&--&--&--&--&--&--
      & --&--&--&--&--&--&--&--\\
      & Qwen
      & \wrong{21}&\wrong{8}&\wrong{9}&\wrong{29}&\wrong{74}&\wrong{74}&\wrong{74}&\wrong{74}&\wrong{74}
      & \wrong{74}&\wrong{74}&\wrong{74}&\wrong{74}&\wrong{74}&\wrong{74}&\wrong{74}&\wrong{7}
      & \wrong{74}&\wrong{74}&\wrong{74}&\wrong{74}&\wrong{38}&\wrong{43}&\wrong{74}&\wrong{95}\\
      & GPT
      & \correct{12}&\correct{3}&\wrong{3}&\wrong{29}&\wrong{55}&\wrong{3}&\correct{5}&\wrong{15}&\wrong{4}
      & \wrong{9}&\wrong{3}&\wrong{37}&\wrong{45}&\textbf{N/A}&\wrong{7}&\wrong{16}&\wrong{13}
      & \wrong{74}&\correct{2}&\wrong{74}&\wrong{27}&\wrong{25}&\textbf{N/A}&\wrong{39}&\wrong{96}\\
    \midrule
    \multirow{8}{*}{Deuteranopia}
      & \cellcolor{blue!10}\textbf{Gold}
      & \textbf{12}&\textbf{3}&\textbf{5}&\textbf{70}&\textbf{35}&\textbf{2}&\textbf{5}&\textbf{17}&\textbf{21}
               &\textbf{N/A}&\textbf{N/A}&\textbf{N/A}&\textbf{N/A}&\textbf{N/A}&\textbf{N/A}&\textbf{N/A}&\textbf{N/A}
               &\textbf{5}&\textbf{2}&\textbf{45}&\textbf{73}&\textbf{2}&\textbf{4}&\textbf{3}&\textbf{9}\\
    \cmidrule(lr){2-27}
    & Llama  & \wrong{19}&\wrong{2}&\correct{5}&\wrong{52}&\wrong{1}&\wrong{53}&\wrong{52}&\wrong{49}&\wrong{45}   
             &\wrong{53}&\wrong{52}&\wrong{52}&\wrong{36}&\wrong{55}&\wrong{7}&\wrong{56}&\wrong{4}   
             &\wrong{4}&\correct{2}&\wrong{49}&\wrong{53}&\wrong{42}&\wrong{52}&\wrong{37}&\wrong{8}\\
    & LLaVA  & \correct{12}&\wrong{8}&\wrong{6}&\wrong{29}&\wrong{57}&\wrong{3}&\wrong{3}&\wrong{16}&\wrong{14}     
             &\wrong{2}&\wrong{6}&\wrong{9}&\wrong{10}&\wrong{6}&\wrong{7}&\wrong{16}&\wrong{3}   
             &\wrong{10}&\wrong{1}&\wrong{1}&\wrong{12}&\wrong{20}&\wrong{42}&\wrong{38}&\wrong{96}\\
    & mPLUG  & \correct{12}&\wrong{8}&\wrong{6}&\wrong{29}&\wrong{37}&\wrong{5}&\wrong{3}&\wrong{15}&\wrong{24}    
             &\wrong{2}&\wrong{6}&\wrong{9}&\wrong{10}&\wrong{5}&\wrong{7}&\wrong{16}&\wrong{13} 
             &\wrong{1}&\wrong{1}&\wrong{10}&\wrong{1}&\wrong{23}&\wrong{42}&\wrong{35}&\wrong{96}\\
    & Phi    & --&--&--&--&--&--&--&--&--        
             &--&--&--&--&--&--&--&--    
             &--&--&--&--&--&--&--&--\\
    & Qwen   & \wrong{21}&\wrong{8}&\wrong{9}&\wrong{29}&\wrong{74}&\wrong{74}&\wrong{74}&\wrong{74}&\wrong{74} 
             &\wrong{74}&\wrong{74}&\wrong{74}&\wrong{74}&\wrong{74}&\wrong{74}&\wrong{53}&\wrong{7}  
             &\wrong{74}&\wrong{74}&\wrong{74}&\wrong{74}&\wrong{88}&\wrong{43}&\wrong{30}&\wrong{95}\\
    & GPT    & \correct{12}&\wrong{8}&\wrong{6}&\wrong{29}&\wrong{57}&\wrong{3}&\wrong{8}&\textbf{N/A}&\wrong{74} 
             &\wrong{2}&\wrong{6}&\wrong{97}&\wrong{45}&\wrong{3}&\wrong{7}&\wrong{16}&\wrong{73} 
             &\wrong{74}&\textbf{N/A}&\textbf{N/A}&\wrong{5}&\wrong{93}&\wrong{42}&\wrong{35}&\wrong{96}\\
    \midrule
    \multirow{8}{*}{Tritanopia}
      & \cellcolor{blue!10}\textbf{Gold}
      & \textbf{12}&\textbf{8}&\textbf{6}&\textbf{29}&\textbf{57}&\textbf{5}&\textbf{3}&\textbf{15}&\textbf{74}
               &\textbf{2}&\textbf{6}&\textbf{97}&\textbf{45}&\textbf{5}&\textbf{7}&\textbf{16}&\textbf{73}
               &\textbf{N/A}&\textbf{N/A}&\textbf{N/A}&\textbf{N/A}&\textbf{26}&\textbf{42}&\textbf{35}&\textbf{96}\\
    \cmidrule(lr){2-27}
    & Llama  & \wrong{19}&\wrong{2}&\wrong{5}&\wrong{17}&\wrong{1}&\wrong{3}&\wrong{2}&\wrong{4}&\wrong{1}   
             &\wrong{3}&\wrong{4}&\wrong{42}&\wrong{46}&\wrong{4}&\correct{7}&\wrong{3}&\wrong{8}   
             &\wrong{4}&\wrong{4}&\wrong{4}&\wrong{1}&\wrong{3}&\wrong{4}&\wrong{3}&\wrong{8}\\
    & LLaVA  & \correct{12}&\correct{8}&\correct{6}&\correct{29}&\wrong{37}&\wrong{3}&\correct{3}&\wrong{16}&\wrong{7}   
             &\correct{2}&\correct{6}&\wrong{9}&\wrong{1}&\wrong{6}&\correct{7}&\correct{16}&\wrong{3}   
             &\wrong{1}&\wrong{1}&\wrong{1}&\wrong{12}&\wrong{28}&\correct{42}&\wrong{38}&\correct{96}\\
    & mPLUG  & \correct{12}&\correct{8}&\correct{6}&\correct{29}&\wrong{37}&\correct{5}&\correct{3}&\correct{15}&\wrong{24} 
             &\wrong{9}&\correct{6}&\wrong{9}&\wrong{10}&\correct{5}&\correct{7}&\correct{16}&\wrong{13} 
             &\wrong{10}&\wrong{10}&\wrong{10}&\wrong{10}&\wrong{23}&\correct{42}&\correct{35}&\correct{96}\\
    & Phi    & --&--&--&--&--&--&--&--&--      
             &--&--&--&--&--&--&--&--    
             &--&--&--&--&--&--&--&--\\
    & Qwen   & \wrong{21}&\correct{8}&\correct{6}&\correct{29}&\wrong{37}&\wrong{74}&\wrong{74}&\wrong{74}&\correct{74} 
             &\wrong{74}&\wrong{74}&\wrong{74}&\wrong{74}&\correct{5}&\wrong{3}&\wrong{53}&\wrong{3}   
             &\wrong{74}&\wrong{0}&\wrong{74}&\wrong{74}&\wrong{38}&\wrong{43}&\wrong{30}&\wrong{95}\\
    & GPT    & \correct{12}&\correct{8}&\correct{6}&\correct{29}&\correct{57}&\textbf{N/A}&\correct{3}&\correct{15}&\correct{74} 
             &\correct{2}&\wrong{3}&\wrong{37}&\wrong{36}&\correct{5}&\correct{7}&\correct{16}&\wrong{33} 
             &\wrong{74}&\wrong{74}&\wrong{6}&\wrong{3}&\textbf{N/A}&\correct{42}&\correct{35}&\correct{96}\\
    \bottomrule
  \end{tabular}%
}
\caption{Predicted outputs of six LVLMs on Ishihara color-vision test plates under four simulated conditions (Normal, Protanopia, Deuteranopia, and Tritanopia) using the Base prompt only. 
The ``Gold'' row shows the ground-truth digits. 
Plates marked as ``N/A'' indicate those without a correct digit (e.g., blank or pattern-only plates) and are excluded from accuracy calculations.}
\label{tab:output_digit_base_all}

\end{table*}

\section{Results and Analysis}

\subsection{RQ1: Generation Level}
\label{sec:rq1}

We use \emph{Digit Accuracy} to evaluate how often the model correctly identifies the numeral on each Ishihara plate. Table~\ref{tab:result-score} shows accuracy across conditions and prompt types. In the Normal condition, humans are expected to score 100 \%, but only GPT-4o approaches this with 90.5 \%, followed by mPLUG-Owl3 at 71 \%, while some models score near 0 \%. Performance collapses for both Protanopia and Deuteranopia, with no model exceeding 24 \%, clearly indicating the difficulty of simulating CVDs. With Linguistic Support, brief descriptions shift accuracy by at most two points without changing rankings; GPT-4o improves to 23 \% on Deuteranopia but drops on Tritanopia. Qwen2.5-VL often refuses under disability-related terminology, averaging 0 \%. With Visual Support, few-shot examples have minimal effect; accuracy shows no consistent improvement and sometimes decreases.

Table~\ref{tab:output_digit_base_all} lists the gold digits and Base-prompt predictions across 25 plates. GPT-4o reproduces most digits under Normal vision, and mPLUG-Owl3 performs well, but the remaining models are inconsistent. Under Protanopia and Deuteranopia, most models repeat Normal outputs, return empty strings, or collapse to a single digit, e.g., ``70.'' Thus, strong Normal accuracy does not imply successful CVD simulation. Even with Linguistic or Visual Support, accuracy improves little and sometimes introduces new errors, indicating that prompt modifications alone cannot reproduce perceptual variation.
Tritanopia shares the same ground-truth digits as Normal and therefore serves as a control condition. Models output nearly identical digits for both, matching clinical answers but indicating fallback to Normal rather than simulation. Severe failures under Protanopia and Deuteranopia further show that agreement in Tritanopia does not imply perceptual simulation.

\emph{At the generation level}, current LVLMs assume normal color vision and, even with prompt support, fail to generate text that matches CVDs perception.

\begin{table*}[t]
  \centering
  \small
  \resizebox{1.00\textwidth}{!}{%
    \begin{tabular}{l r l rr rr rr rr}
      \toprule
      \multirow{2.25}{*}{\textbf{LVLM}} & \multirow{2.25}{*}{\textbf{Size}} & \multirow{2.25}{*}{\textbf{Prompt}} 
        & \multicolumn{2}{c}{\textbf{Normal}} 
        & \multicolumn{2}{c}{\textbf{Protanopia}} 
        & \multicolumn{2}{c}{\textbf{Deuteranopia}} 
        & \multicolumn{2}{c}{\textbf{Tritanopia}} \\
        \cmidrule(lr){4-5}\cmidrule(lr){6-7}\cmidrule(lr){8-9}\cmidrule(lr){10-11}
      & & 
        & \textbf{Mean} & \textbf{SD}
        & \textbf{Mean} & \textbf{SD}
        & \textbf{Mean} & \textbf{SD}
        & \textbf{Mean} & \textbf{SD} \\
      \midrule
      \multirow{3}{*}{mPLUG‑Owl3}  & \multirow{3}{*}{7B}
        & Base       & 3.8 & 6.7 & 69.7 & 78.8 & 69.0 & 116.0 & 3.7 & 6.3 \\
      & & Linguistic & 3.8 & 6.7 & 78.9 & 101.5 & 105.9 & 185.6 & 4.2 & 6.9 \\
      & & Visual     & 16.0 & 40.2 & 36.5 & 43.3 & 37.1 & 59.4 & 14.3 & 33.5 \\
      \midrule
      \multirow{3}{*}{Phi‑3.5}     & \multirow{3}{*}{4.2B}
        & Base       & 14.5 & 13.3 & 140.5 & 325.5 & 80.0 & 256.1 & 23.9 & 28.6 \\
      & & Linguistic & 14.5 & 13.3 & 334.0 & 792.7 & 158.2 & 544.1 & 24.9 & 28.7 \\
      & & Visual     & 7.0 & 6.2 & 13.1 & 12.3 & 12.7 & 12.8 & 7.6 & 7.3 \\
      \midrule
      \multirow{3}{*}{Qwen2.5‑VL}   & \multirow{3}{*}{7B}
        & Base       & 46.0 & 83.3 & 72.8 & 62.1 & 57.2 & 69.1 & 27.4 & 36.6 \\
      & & Linguistic & 46.0 & 83.3 & 62.0 & 68.7 & 51.4 & 59.7 & 27.7 & 36.6 \\
      & & Visual     & 45.0 & 56.5 & 313.7 & 381.8 & 131.7 & 172.0 & 78.0 & 82.1 \\
      \bottomrule
    \end{tabular}%
  }
  \caption{Mean and standard deviation (SD) of \emph{per-token perplexity} across four simulated vision conditions (Normal, Protanopia, Deuteranopia, Tritanopia) under three prompt settings (Base, Linguistic, Visual) for each LVLM. 
Only models that could be tested under all three settings are shown. }
  
  \label{tab:result-ppl-vertical}
\end{table*}

\begin{figure*}[ht]
\centering
 \includegraphics[width=1.00\textwidth]{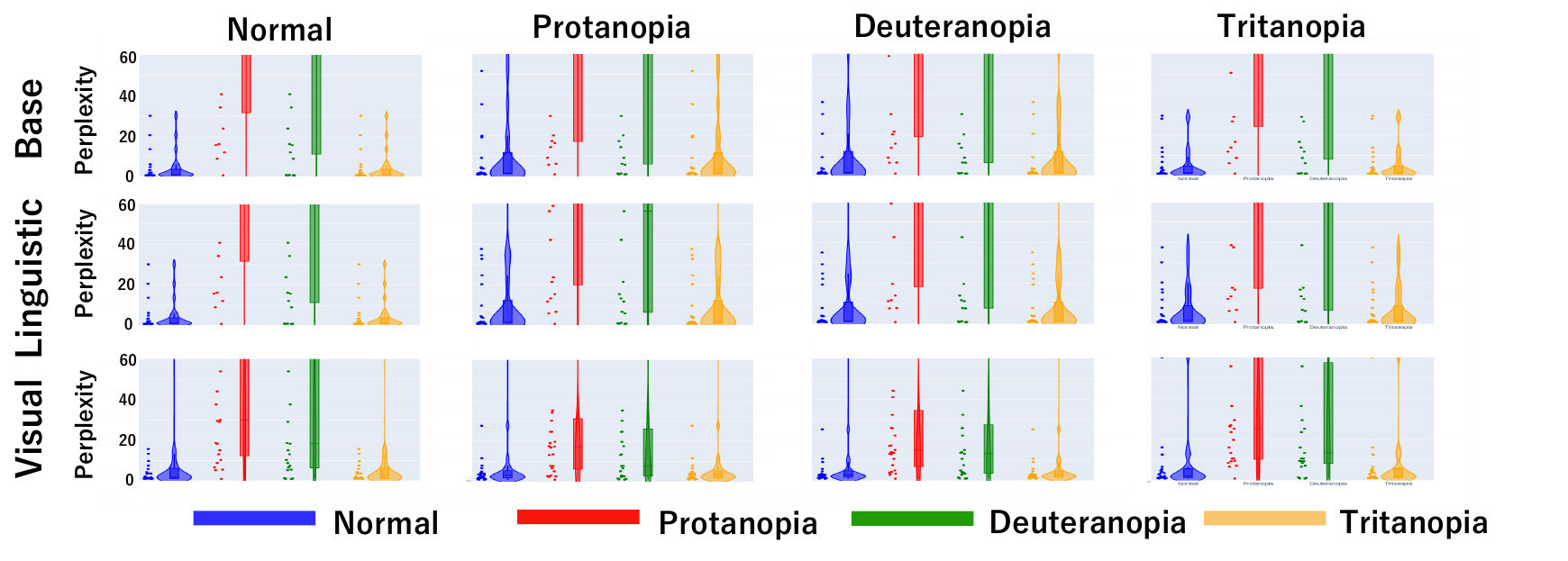}
\caption{Violin plots of mPLUG-Owl3 perplexity on Ishihara digit prediction. Columns indicate the condition specified in the prompt (Normal, Protanopia, Deuteranopia, Tritanopia). Within each column, colors denote the vision type that defines the candidate digit whose perplexity is measured (Normal: blue, Protanopia: red, Deuteranopia: green, Tritanopia: yellow). Lower perplexity (for the gold answer) indicates higher model confidence.}
\label{fig:dataflow}
\end{figure*}

\subsection{RQ2: Confidence Level}
\label{sec:rq2}
To evaluate model confidence, we measure \emph{per-token perplexity} for each forced-decoded output.
Comparing perplexity across conditions reveals whether confidence aligns with accuracy, whether linguistic or visual cues reduce uncertainty, and how strongly models default to normal vision.

Table~\ref{tab:result-ppl-vertical} reports the mean and standard deviation (SD) of perplexity (shown as mean~$\pm$~SD) for all models under the three prompt settings.
Across the board, Normal and Tritanopia remain lower than Protanopia and Deuteranopia, indicating a pronounced bias toward normal vision, though exceptions are observed in Llama-3.2.
One model rises from about $4 \pm 7$ on Normal to $70 \pm 80$ on Protanopia, while another climbs from about $14 \pm 13$ to $140 \pm 330$.
Perplexity generally increases under Linguistic and Visual Support. Although visual cues lower perplexity for a few models, the tendency persists: Normal is always lowest and the red and green deficiencies remain highest.

\begin{table}[t]
  \centering
  \small
  \renewcommand{\arraystretch}{1.00} 
  \resizebox{1.00\columnwidth}{!}{%
    \begin{tabular}{l r r r}
      \toprule
      \textbf{LVLM} & \textbf{Size} & \textbf{Normal} & \textbf{Protanopia} \\
      \midrule
      Llama-3.2   & 11B  & 36.0  & 72.4   \\
      LLaVA-NeXT   & 13B  & 20.0  & 86.2  \\
      mPLUG-Owl3   & 7B   & 28.0  & 86.2  \\
      Phi-3.5     & 4.2B & 100.0   & 3.5   \\
      Qwen2.5-VL    & 7B   & 4.0  & 86.2   \\
      \bottomrule
    \end{tabular}%
  }
  \caption{Doctor-style diagnosis accuracy (\%) under Normal and Protanopia conditions for various LVLMs.}
  \label{tab:result-score-doctor}
\end{table}

\begin{figure}[ht]
\centering
 \includegraphics[width=\columnwidth]{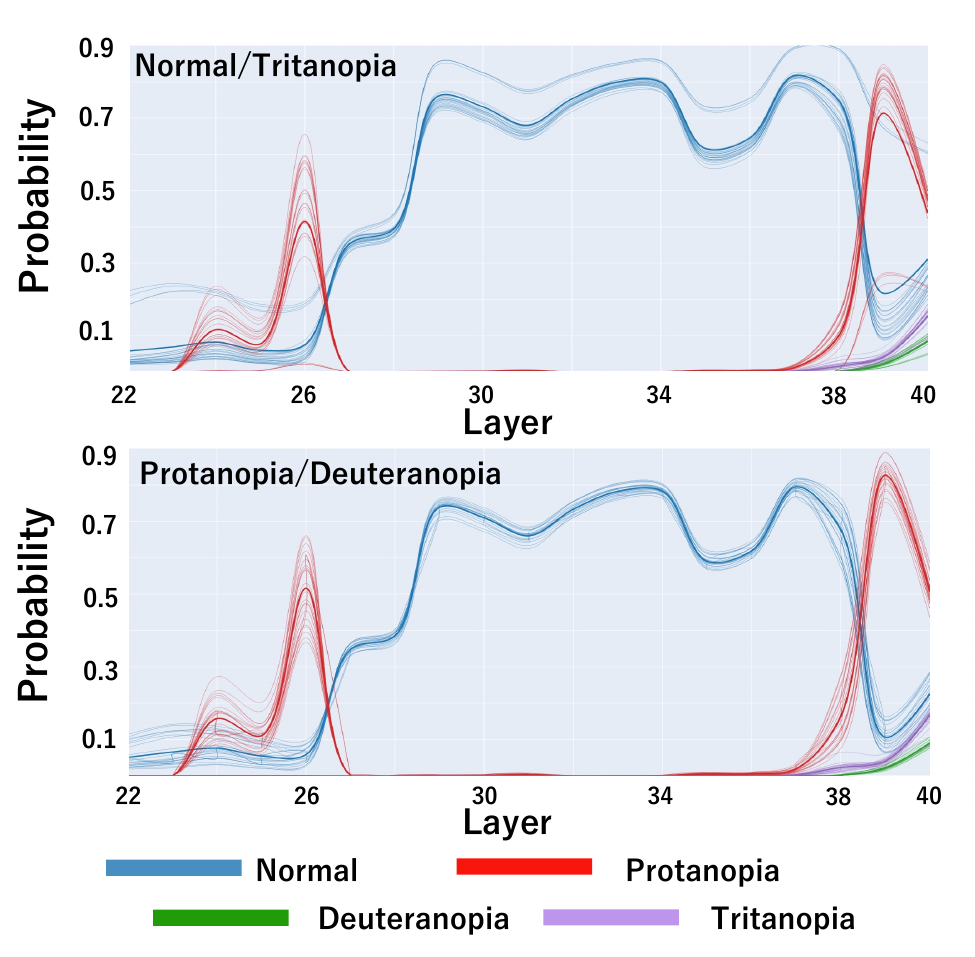}
\caption{\emph{Layer-wise diagnosis probabilities} for LLaVA, averaged over all plates. }

\label{fig:lense_llava-main}
\end{figure}

\begin{table*}[t]
  \centering
  \resizebox{1.00\textwidth}{!}{%
    \begin{tabular}{l r  rrr | rrr | rrr | rrr | rrr}
      \toprule
       & 
        & \multicolumn{3}{c}{\textbf{Normal}}
        & \multicolumn{3}{c}{\textbf{Protanopia}}
        & \multicolumn{3}{c}{\textbf{Deuteranopia}}
        & \multicolumn{3}{c}{\textbf{Tritanopia}}
        & \multicolumn{3}{c}{\textbf{Avg.}} \\
      \cmidrule(lr){3-5} \cmidrule(lr){6-8} \cmidrule(lr){9-11} \cmidrule(lr){12-14} \cmidrule(lr){15-17}
        \multirow{-2.25}{*}{\textbf{LVLM}} &  \multirow{-2.25}{*}{\textbf{Size}}
        & \textbf{Base } & \textbf{Ling.} & \textbf{Vis. }
        & \textbf{Base } & \textbf{Ling. } & \textbf{Vis. }
        & \textbf{Base } & \textbf{Ling. } & \textbf{Vis. }
        & \textbf{Base } & \textbf{Ling. } & \textbf{Vis. }
        & \textbf{Base } & \textbf{Ling.} & \textbf{Vis. } \\
      \midrule
      mPLUG-Owl3   & 7B   & 71.4 & 71.4 & 62.0 & 5.9 & 5.9 & 5.9 & 5.9 & 5.9 & 17.7 & 66.7 & 66.7 & 62.0 & 37.5 & 37.5 & 36.9 \\
      Qwen2.5-VL   & 7B   & 28.6 & 0.0 & 4.8 & 0.0 & 0.0 & 5.9 & 0.0 & 0.0 & 0.0 & 23.8 & 0.0 & 4.8 & 13.1 & 0.0 & 3.9 \\
      GPT-4o       & --   & 90.5 & 90.5 & 90.5 & 23.6 & 17.7 & 23.6 & 5.9 & 23.5 & 5.9 & 71.4 & 61.9 & 42.9 & 47.9 & 48.4 & 40.7 \\
      \midrule
      MedVLM-R1     & --   & 57.1 & 57.1 & 0.0 & 5.9 & 5.9 & 0.0 & 5.9 & 5.9 & 0.0 & 61.9 & 61.9 & 0.0 & 32.7 & 32.7 & 0.0 \\
      Qwen‑DotPattern    & --   & 71.4 & 71.4 & 23.8 & 5.9 & 5.9 & 5.9 & 5.9 & 5.9 & 17.7 & 66.7 & 66.6 & 23.8 & 37.5 & 37.5 & 17.8 \\
      \bottomrule
    \end{tabular}%
  }
\caption{\emph{Digit accuracy} (\%) of each LVLM under simulated vision types: \textbf{Normal}, \textbf{Protanopia}, \textbf{Deuteranopia}, and \textbf{Tritanopia}, with support settings: \textbf{Base}, \textbf{Ling.}, and \textbf{Vis.} 
As for \textbf{Vis.}, it was used only for models supporting multi-image input. 
This table extends Table~\ref{tab:result-score} by including additional results for MedVLM-R1 and Qwen-DotPattern.}
\label{tab:medvlm_results}
\end{table*}

Figure~\ref{fig:dataflow} plots perplexity for mPLUG-Owl3. We selected this model because it supports multiple images and exhibits the most stable perplexity across prompt settings.  The three prompt settings share nearly identical shapes; Normal clusters near zero, whereas Protanopia and Deuteranopia rise to tall peaks. Adding a brief description or a reference image leaves the distribution almost unchanged, which shows that prompt tweaks do not alter the model's internal uncertainty.
Other results are in Appendix~\ref{appendix:perplexity_results}, showing the same trend: low perplexity for Normal vision, higher and variable for CVDs, with little effect from prompt changes.

\emph{At the confidence level}, Normal vision consistently yields the lowest perplexity, while the CVD conditions result in similarly high perplexity, suggesting that models are fundamentally confused when simulating altered color vision and fail to reproduce such perceptual states.

\subsection{RQ3: Internal Representation Level}
\label{sec:rq3}
We analyze whether LVLMs can perform doctor-style diagnosis by inferring vision types from responses using the Doctor-Style prompt (Table~\ref{tab:representative_prompts}).
This setting evaluates whether models can identify vision types from a response, analogous to how a doctor interprets Ishihara results.
For each plate, we force-decode four labels (Normal, Protanopia, Deuteranopia, and Tritanopia) and select the most probable one.
As shown in Table~\ref{tab:result-score-doctor}, models perform well on Protanopia (over 70\%, except Phi-3.5), but poorly on Normal (under 30\%).  
This contrasts with Table~\ref{tab:result-score}, where they failed to recognize digits under red and green conditions.  

\begin{table*}[t]
  \centering
  \small
  \resizebox{1.00\textwidth}{!}{%
    \begin{tabular}{l r l rr rr rr rr}
      \toprule
      \multirow{2.25}{*}{\textbf{LVLM}} & \multirow{2.25}{*}{\textbf{Size}} & \multirow{2.25}{*}{\textbf{Prompt}} 
        & \multicolumn{2}{c}{\textbf{Normal}} 
        & \multicolumn{2}{c}{\textbf{Protanopia}} 
        & \multicolumn{2}{c}{\textbf{Deuteranopia}} 
        & \multicolumn{2}{c}{\textbf{Tritanopia}} \\
        \cmidrule(lr){4-5}\cmidrule(lr){6-7}\cmidrule(lr){8-9}\cmidrule(lr){10-11}
      & & 
        & \textbf{Mean} & \textbf{SD} 
        & \textbf{Mean} & \textbf{SD} 
        & \textbf{Mean} & \textbf{SD} 
        & \textbf{Mean} & \textbf{SD} \\
      \midrule
      \multirow{3}{*}{Qwen2.5‑VL}   & \multirow{3}{*}{7B}
        & Base       & 46.0 & 83.3 & 72.8 & 62.1 & 57.2 & 69.1 & 27.4 & 36.6 \\
      & & Linguistic & 46.0 & 83.3 & 62.0 & 68.7 & 51.4 & 59.7 & 27.7 & 36.6 \\
      & & Visual     & 45.0 & 56.5 & 313.7 & 381.8 & 131.7 & 172.0 & 78.0 & 82.1 \\
      \midrule
      \multirow{3}{*}{MedVLM-R1}   & \multirow{3}{*}{2B}
        & Base       & 15.5 & 28.6 & 166.4 & 325.3 & 86.1 & 160.8 & 14.7 & 26.2 \\
      & & Linguistic & 15.5 & 28.6 & 189.7 & 369.6 & 91.5 & 195.2 & 13.2 & 24.5 \\
      & & Visual     & 48.6 & 63.6 & 83.7 & 92.3 & 72.7 & 89.6 & 50.9 & 62.1 \\
      \midrule
      \multirow{3}{*}{Qwen‑DotPattern}   & \multirow{3}{*}{2B}
        & Base       & 4.7 & 8.9 & 79.2 & 136.3 & 75.4 & 171.7 & 4.7 & 6.6 \\
      & & Linguistic & 4.7 & 8.9 & 106.3 & 213.0 & 100.2 & 249.9 & 4.7 & 6.5 \\
      & & Visual     & 6.2 & 9.1 & 41.5 & 55.2 & 28.7 & 47.1 & 7.4 & 9.9 \\
      \bottomrule
    \end{tabular}%
  }
\caption{Mean and standard deviation (SD) of \emph{per-token perplexity} across four simulated vision conditions under three prompt settings (Base, Linguistic, Visual). 
This table extends Table~\ref{tab:result-ppl-vertical} by including additional results for MedVLM-R1 and Qwen-DotPattern.}

  \label{tab:result-ppl-domain}
\end{table*}

\begin{table}[t]
  \centering
  \small
  \renewcommand{\arraystretch}{1.00} 
  \resizebox{1.00\columnwidth}{!}{%
    \begin{tabular}{l c r r}
      \toprule
      \textbf{LVLM} & \textbf{Size} & \textbf{Normal} & \textbf{Protanopia} \\
      \midrule
      Qwen2.5-VL    & 7B   & 4.0  & 86.2   \\
      MedVLM-R1     & 7B   & 21.4  & 88.0   \\
      Qwen-DotPattern  & 7B   & 85.7 & 0.0   \\
      \bottomrule
    \end{tabular}%
  }
\caption{Doctor-style diagnosis accuracy (\%) under Normal and Protanopia conditions for MedVLM-R1 and Qwen-DotPattern, extending Table~\ref{tab:result-score-doctor}.}
  \label{tab:result-score-doctor-domain}
\end{table}

To further examine this behavior, we use \emph{Layer-Wise Diagnosis Probability}, inspired by Logit Lens~\cite{nostalgebraist2020logitlens}, to track token-level probabilities across transformer layers. 
Figure~\ref{fig:lense_llava-main} uses LLaVA-NeXT as a representative example.  
Results for other models are provided in Appendix~\ref{appendix:doctor_layers_all},  
where no consistent trends were observed across models, indicating that the layer-wise behavior varies depending on the architecture.  
LLaVA-NeXT tends to interpret images as Normal in mid-layers, shifting to Protanopia at the final layer, suggesting inconsistent processing.  
Nearly identical curves across ground-truths imply poor differentiation between conditions. 

\emph{At the internal representation level}, the model exhibits nearly identical behavior across all color vision types, showing no evidence of reasoning that reflects differences between conditions. This indicates a fundamental inability to understand and reproduce variations in color perception.

\subsection{Domain Knowledge and Pattern Memorization}
\label{sec:medvlm_dot}

To deepen our analysis, we conducted two focused experiments as outlined below.
\begin{description}
  \item[Effect of medical expertise] 
        As Ishihara plates originate in clinical practice, we expect that an LVLM trained on medical images and reports would outperform standard models.
        We therefore evaluated MedVLM‑R1 \cite{pan2025medvlmr1incentivizingmedicalreasoning}, which is built on Qwen‑2.0‑VL \cite{wang2024qwen2} and fine‑tuned on medical data, to investigate whether training on clinical data improves performance on color-based diagnostic tools.

  \item [Dot pattern memorization hypothesis]  
        A potential concern is that models may solve the task by exploiting superficial dot-pattern regularities in Ishihara plates, rather than demonstrating perceptual reasoning.
        To directly probe this, we fine‑tuned Qwen2.0‑VL on 10,000 synthetic Ishihara‑style images released by \citet{grover2025huemanityprobingfinegrainedvisual}, producing a model specialized in dot‑pattern recognition.
        We refer to this model as Qwen‑DotPattern.
        This setup allows us to assess whether pattern memorization alone can explain model performance.
        Training details are provided in Appendix~\ref{sec:appendix:train-settings}.

\end{description}

Table~\ref{tab:medvlm_results} shows generation-level accuracy for MedVLM-R1 and Qwen-DotPattern. 
Under the Normal condition with the Base prompt, both outperform the Qwen2.5-VL baseline (MedVLM-R1: 57.1\%, Qwen-DotPattern: 71.4\%), indicating that domain-specific training with dot-pattern stimuli enhances recognition under typical vision. 
However, under Protanopia and Deuteranopia, follow the same trend as other models, showing low accuracy across prompts, suggesting that neither medical knowledge nor pattern memorization reproduces CVD perceptual distortions.

Table~\ref{tab:result-ppl-domain} presents the confidence level results in terms of perplexity.
Under the Normal condition, both MedVLM-R1 and Qwen-DotPattern show lower perplexity than Qwen2.5-VL, indicating that domain adaptation enhances confidence on typical inputs.
However, under red–green CVDs, both show higher perplexity than the base model, suggesting that medical or synthetic training improves confidence in normal vision but increases miscalibration when color perception is impaired.

We extend the doctor diagnosis analysis by adding MedVLM-R1 and Qwen-DotPattern, as shown in Table~\ref{tab:result-score-doctor-domain}.
MedVLM-R1 shows similar accuracy in both Normal (21.4\%) and Protanopia (88.0\%) settings, consistent with general model trends.
This suggests that while medical training supports classification under impaired vision, it does not improve Normal vision.
Qwen-DotPattern shows the opposite pattern.
It performs well under Normal (85.7\%) but fails under Protanopia (0.0 \%).
This suggests the model may rely on memorized dot layouts rather than learning generalizable structure, improving recognition in standard settings but reducing adaptability under altered vision.

These results show that training on domain-specific and dot-pattern images can improve model behavior in normal color vision, but do not lead to gains in reproducing the perceptual experience of CVDs.
This highlights the value of the Ishihara Test, which remains a strong diagnostic benchmark despite its small size, as it is unaffected by pattern memorization and reliably evaluates a model's reasoning under impaired color perception.

\section{Conclusion}
\label{sec:conclusion}
This work examined whether LVLMs can simulate variation in human color perception using the Ishihara Test. We analyzed model behavior at the levels of generation, confidence, and internal representation to assess whether linguistic knowledge about CVDs reflects perceptual grounding.
Although LVLMs can describe CVDs and the test procedure, they default to a normative color percept and do not reproduce the digit responses associated with different vision types. Additional textual guidance or visual manipulation does not induce alternative percepts, and probing reveals only weak separation between vision types in internal activations.
These findings indicate that current LVLMs lack mechanisms for alternative perceptual experiences, highlighting the need for approaches that support perceptual diversity in future multimodal systems for accessible and human-centered deployment.

\section{Limitations}
\label{sec:limitations}

\paragraph{Dataset Scope and Scale.}
Our evaluation relies on a small set of Ishihara plates that primarily target red–green color-vision deficiencies. The limited number and narrow focus reduce the diversity of conditions tested, so our findings may not generalise to other CVD types or broader visual tasks. While scaling to broader datasets and real-world tasks remains important future work, our focus here is on establishing a methodology and analysing behavioural differences across prompted vision conditions rather than memorisation. Possible pre-training exposure to Ishihara plates would mainly affect the Normal condition and does not account for failures under CVD prompts.

\paragraph{Limitations in Evaluating Tritanopia.}
The Ishihara Test is primarily designed to detect red–green deficiencies (Protanopia and Deuteranopia) and is not suitable for assessing blue–yellow deficiencies such as Tritanopia. Moreover, the medically defined ground-truth answers for Tritanopia coincide with those for Normal vision. In our setting, Tritanopia therefore functions largely as a control for instruction following rather than as a genuine benchmark for blue–yellow deficiencies. As a result, conclusions on Tritanopia are limited, and future work should evaluate all major CVD types using more general tasks beyond Ishihara plates.

\paragraph{Evaluation Methodology.}
We prompt LVLMs to simulate color-blind perception in a controlled diagnostic setting, tasking them with identifying digits in Ishihara plates. We use a restricted set of prompts (role-playing, doctor-perspective, step-by-step). While failures were broadly consistent across them, the prompt space is incomplete and other formulations may elicit different behaviours. Consequently, our conclusions characterise LVLM behaviour under this evaluation format rather than ruling out prompts that could induce more faithful CVD simulation. We also treat medically defined Ishihara digits as ground truth without validation from individuals with CVD or clinicians, which may limit ecological validity.

\paragraph{Generalisability of Model Behavior.}
Performance on stylised Ishihara plates does not imply that a model can handle color perception in broader settings. Models may rely on dot-pattern heuristics or memorised cues rather than simulating color confusion. Prior work also shows that vision–language models can exhibit color biases~\cite{raj-etal-2024-biasdora}, suggesting that controlled diagnostic performance may not translate to broader color understanding. Caution is therefore required when extrapolating to natural images. Finally, this study is diagnostic rather than prescriptive: we do not test mitigation strategies, and developing such methods remains an important direction for CVD-aware LVLMs.

\section{Ethical Considerations}
\label{sec:ethics}

\paragraph{Inclusive Design for Perceptual Diversity.}  
Recent AI research communities, including ACL, have emphasized the importance of designing fair and inclusive systems that accommodate users with perceptual or cognitive differences.  
Color vision deficiency, which affects a significant portion of the global population, is one such factor that is often overlooked in model development and evaluation.  
By explicitly testing LVLM behavior under CVD conditions, this study contributes to a broader understanding of accessibility in multimodal AI systems and encourages further research in this direction.  
Building on this motivation, we also discuss fairness-related concerns in the context of model evaluation.

\paragraph{Not for Diagnostic Use.}
This study aims to evaluate whether models can simulate aspects of color vision deficiency as a step toward developing systems that better capture the diversity of human perception.  
It is not intended to replace professional diagnosis or support clinical decision-making.  
Even if a model produces correct responses to certain vision tests, these outputs must not be used for medical purposes or institutional assessments.  
To prevent such misuse, we clearly define the scope and limitations of our evaluation and explicitly state that the outputs are not a substitute for expert judgment in clinical contexts.

\paragraph{Licensing and Copyright.}
The Ishihara Test \cite{ishihara1917test} is a long-established diagnostic tool widely used in academic research across fields such as ophthalmology, psychology, education, and machine learning~\citep{zhao2019comparison, klinke2024comparative,  s19102250}.
We did not reproduce, modify, or redistribute the plates; they were used solely for research and analysis.

\paragraph{AI-assistant Usage.}
Portions of the manuscript, such as prompt templates and wording adjustments, were drafted with the assistance of GPT-4 to ensure linguistic clarity.  
All technical content, analysis scripts, and final decisions were made by us.

\bibliography{custom}

\appendix

\section{Detailed Model Settings}
\label{appendix:Detailed Settings}
In this study, to ensure fair and consistent performance comparisons, all experiments were conducted using a single NVIDIA RTX 6000 Ada GPU.
All models were run under the same computational settings, with generation performed using half quantization to reduce memory usage while maintaining model fidelity.
We used publicly available models from Hugging Face and OpenAI, as summarized in Table~\ref{tab:detailed-model-names}.
Generation was conducted using greedy decoding, with fixed random seeds to ensure reproducibility.
All other settings adhered to the default configurations of each model implementation.

\begin{table}[ht!]
    \centering
     \resizebox{\linewidth}{!}{%
    \begin{tabular}{lcl}
        \toprule
        \textbf{Model}  & \textbf{Params} & \textbf{HuggingFace Name / OpenAI Name}  \\
        \midrule
        \texttt{Phi3.5} & \texttt{4.2B}  & \texttt{microsoft/Phi-3-vision-128k-instruct}  \\
        \texttt{Qwen2.5-VL} & \texttt{7B}  & \texttt{Qwen/Qwen-2.5-VL-7B-Instruct}  \\
        \texttt{mPLUG} & \texttt{7B}  & \texttt{mPLUG/mPLUG-Owl3-7B-240728}  \\
        \texttt{Llama3.2} & \texttt{11B}  & \texttt{meta-llama/Llama-3.2-11B-Vision-Instruct}  \\
        \texttt{LLaVA-NeXT} & \texttt{13B}  & \texttt{llava-hf/llava-v1.6-vicuna-13b-hf}  \\
        \texttt{MedVLM-R1} & \texttt{2B}  & \texttt{JZPeterPan/MedVLM-R1}  \\
        \texttt{GPT-4o} & --  & \texttt{gpt-4o-2024-11-20} \\
        \bottomrule
    \end{tabular}
    }   
    \caption{Detailed model names.}
    \label{tab:detailed-model-names}
\end{table}

\section{Language Knowledge of the Ishihara Test and CVDs}
\label{appendix:cvd_knowledge}
Before testing visual ability, we verified that all LVLMs possess basic language knowledge of the Ishihara Test and the three major color vision deficiencies.
As the main paper includes only one example, this appendix provides the full outputs for the remaining models.

Tables~\ref{tab:llama_cvd_descriptions}--\ref{tab:Qwen_cvd_descriptions} show representative answers from Llama-3.2, LLaVA-NeXT, mPLUG-Owl3, Phi-3.5, and Qwen2.5-VL.
Each table has input prompts on the left and trimmed model responses on the right.
All models except Phi-3.5, which returned an error for the Ishihara query, accurately describe the test and define Protanopia, Deuteranopia, and Tritanopia, confirming sufficient textual knowledge.
These results show that failures in the main experiments arise from weaknesses in visual grounding, not language understanding.

\paragraph{Prompts in Detail}
\label{appendix:prompts_detail}
For each color vision condition, we designed three types of prompts: Base, Linguistic Support, and Visual Support. 
The full prompt templates are provided in Tables~\ref{tab:prompts_base}--\ref{tab:prompts_visual}.

Table~\ref{tab:prompts_base} presents the \textit{Base} prompts, which specify only the condition (e.g., ``You are Protanopic'') and then ask the model to report the number shown in the Ishihara plate. This represents the simplest setting, testing whether the model can directly simulate the specified condition.

Table~\ref{tab:prompts_linguistic} illustrates the \textit{Linguistic Support} prompts, where a short textual description of the impairment (e.g., ``red tones appear dimmer and may be confused with green or brown'') is added. This design evaluates whether providing linguistic context improves the model’s responses.

Table~\ref{tab:prompts_visual} shows the \textit{Visual Support} prompts. In this case, the condition is accompanied by a few shot reference example (e.g., ``A Protanopic person sees this number: \{few\_shot\_number\}'') before asking the model to identify the number in a new plate. This setup tests whether visual exemplars serve as useful guidance for simulating CVD perception.

Together, these tables make explicit how the three prompt types were implemented, complementing the overview provided in Section~\ref{appendix:prompts}.

\section{Effects of Prompt Variants on Simulating CVD Perception}
\label{appendix:prompt_variants}

Prompt phrasing can influence how LVLMs interpret images, particularly when the task requires aligning visual input with an altered perceptual perspective. To assess whether failures in our main experiments were attributable to prompt design rather than perceptual grounding, we evaluated four alternative prompt formulations for the Protanopia condition:

\begin{itemize}
    \item \textbf{Prompt 1 (Description-Based)}:
    ``What number would a Protanopic viewer likely see in this image? Only provide the predicted number.''
    \item \textbf{Prompt 2 (Doctor Perspective)}:
    ``You are an eye doctor evaluating how different color vision types perceive this test plate. What number would a Protanopic patient report? Only output the number.''
    \item \textbf{Prompt 3 (Role-Play Perspective)}:
    ``Answer as if you are a Protanopic viewer. Respond with the number you visually perceive. Only output the number.''
    \item \textbf{Prompt 4 (Step-by-Step Reasoning)}:
    ``Step 1: Describe how the red and green dots differ under Protanopia. Step 2: Based on Step 1, what number appears to a Protanopic viewer? Only output Step~2.''
\end{itemize}

Table~\ref{tab:prompt_variant_digit} reports digit accuracy under Normal and three CVD conditions for Prompts 1--3. Prompt~4 was excluded due to highly unstable behavior: models frequently produced free-form explanations, failed to emit a digit token in Step~2, or ignored the requested output format entirely.

Across all prompt variants, we observe moderate variability in the Normal condition, confirming that prompt phrasing can influence how LVLMs interpret visual input when no perceptual transformation is required. However, under Protanopia, Deuteranopia, and Tritanopia, accuracy remains low across prompts and models. Neither altering task framing (doctor vs. viewer) nor introducing explicit role-playing improved condition-appropriate responses. These results indicate that prompt formulation alone is insufficient to induce LVLMs to simulate condition-specific perceptual distortions, consistent with the main conclusion that limitations arise from insufficient visual grounding rather than linguistic control alone.

\begin{table*}[t]
  \centering
  \small
  \resizebox{1.00\textwidth}{!}{%
    \begin{tabular}{l r  rrr | rrr | rrr | rrr}
      \toprule
       & 
        & \multicolumn{3}{c}{\textbf{Normal}}
        & \multicolumn{3}{c}{\textbf{Protanopia}}
        & \multicolumn{3}{c}{\textbf{Deuteranopia}}
        & \multicolumn{3}{c}{\textbf{Tritanopia}} \\
      \cmidrule(lr){3-5} \cmidrule(lr){6-8} \cmidrule(lr){9-11} \cmidrule(lr){12-14}
      \multirow{-2}{*}{\textbf{LVLM}} & \multirow{-2}{*}{\textbf{Size}}
        & \textbf{P1} & \textbf{P2} & \textbf{P3}
        & \textbf{P1} & \textbf{P2} & \textbf{P3}
        & \textbf{P1} & \textbf{P2} & \textbf{P3}
        & \textbf{P1} & \textbf{P2} & \textbf{P3} \\
      \midrule
      Llama-3.2     & 11B  & 9.5  & 4.8  & 9.5  & 17.6 & 11.8 & 0.0  & 11.8 & 11.8 & 11.8 & 4.8  & 4.8  & 4.8  \\
      LLaVA-NeXT    & 13B  & 61.9 & 52.4 & 57.1 & 5.9  & 5.9  & 5.9  & 5.9  & 5.9  & 5.9  & 47.6 & 61.9 & 57.1 \\
      mPLUG-Owl3    & 7B   & 71.4 & 76.2 & 66.7 & 5.9  & 5.9  & 5.9  & 5.9  & 5.9  & 5.9  & 66.7 & 71.4 & 66.7 \\
      Phi-3.5       & 4.2B & 0.0  & 0.0  & 0.0  & 0.0  & 0.0  & 0.0  & 0.0  & 0.0  & 0.0  & 0.0  & 0.0  & 0.0  \\
      Qwen2.5-VL    & 7B   & 28.6 & 28.6 & 38.1 & 5.9  & 0.0  & 0.0  & 0.0  & 0.0  & 0.0  & 14.3 & 19.1 & 28.6 \\
      \bottomrule
    \end{tabular}
  }
  \caption{\textbf{Digit accuracy} (\%) under three prompt variants (\textbf{P1--P3}) across simulated vision types. Prompt~4 is omitted due to unstable output formatting and inconsistent digit emission.}
  \label{tab:prompt_variant_digit}
\end{table*}

\section{Digit Results: Detailed Analysis}
\label{appendix:digit_results}
Tables~\ref{tab:output_digit_linguistic} and \ref{tab:output_digit_visual} report the raw digit outputs that each LVLM produced for Ishihara plates under four simulated conditions (Normal, Protanopia, Deuteranopia, Tritanopia). Unless stated otherwise, plates marked as ``N/A'' (no correct digit) are excluded from accuracy and are discussed only for error behavior.

\subsection{Linguistic Support prompts (Table~\ref{tab:output_digit_linguistic}).}
Adding a short textual description yields only modest and inconsistent gains. Under Normal vision, GPT remains strong and LLaVA/mPLUG perform similarly to the Base setting, while Llama continues to miss many plates. Under simulated CVDs, condition-specific digits are seldom produced; the same plates that were difficult under Base remain challenging, and response biases (e.g., repeated 74/42) persist. Hallucinated digits also appear on \texttt{N/A} plates, showing that brief definitions alone are insufficient to induce condition-aware perception.

\subsection{Visual Support prompts (Table~\ref{tab:output_digit_visual}).}
Providing a few shot reference image before the target plate produces mixed results. GPT maintains strong accuracy under Normal vision but shows limited and inconsistent improvements for simulated CVDs. mPLUG performs slightly better on some plates compared to Base/Linguistic settings but still fails to reliably follow the instructed condition. Phi occasionally outputs correct answers without cross-plate consistency, while Qwen remains unstable and frequently repeats the same digits. Some regressions and hallucinated digits are also observed, indicating that single visual exemplars do not reliably induce condition-appropriate perception.

\subsection{Overall Findings}
Across all prompt types, LVLMs display recurring error patterns: frequent mode collapse to a small set of digits (e.g., 74, 42, 3), insensitivity to the instructed CVD condition, and hallucinated outputs on plates without valid digits. These behaviors show that neither short textual descriptions nor single visual exemplars are sufficient to ground condition-specific perception. While the models can describe CVDs accurately in text, they fail to translate these instructions into consistent digit recognition on Ishihara plates. This confirms that the main limitation lies in visual grounding rather than language knowledge or prompt design.

\section{Additional Perplexity Analyses}
\label{appendix:perplexity_results}

In the main text, we focused on mPLUG-Owl3 (Figure~\ref{fig:dataflow}) because it supports multi-image input and shows the most stable perplexity across prompt settings. For completeness, we report violin plots of perplexity distributions for three additional models: LLaMA, Phi, and Qwen. 

\begin{figure}[!h]
\centering
 \includegraphics[width=1.00\textwidth]{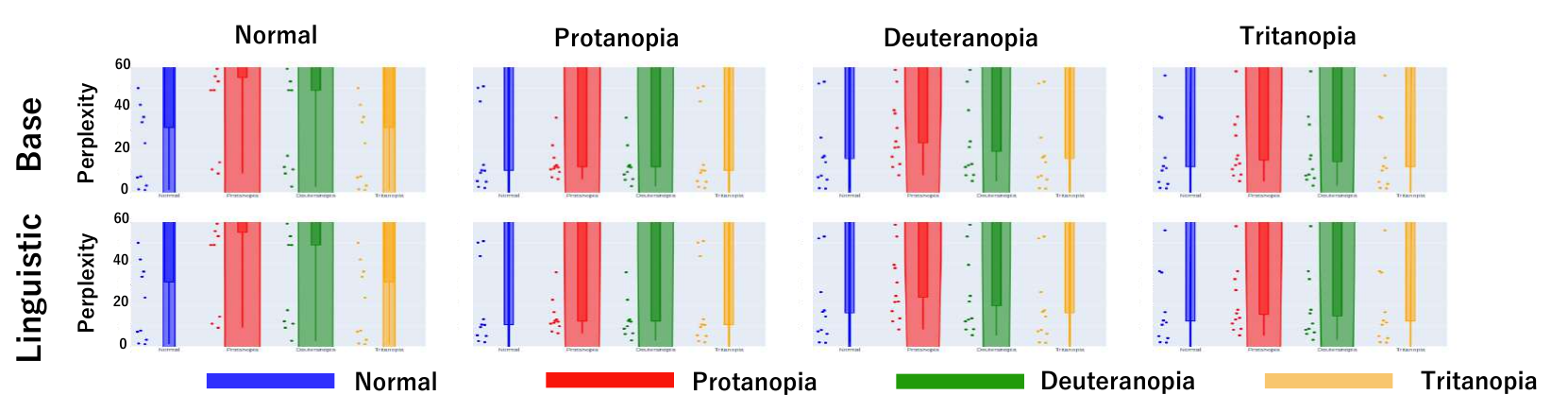}
\caption{Violin plots of LLama perplexity distributions for Ishihara digit predictions under four simulated vision conditions (Normal–blue, Protanopia–red, Deuteranopia–green, Tritanopia–orange) and three prompt types (Base, Linguistic, Visual). For each condition, perplexity is measured by force decoding the Gold Answer corresponding to the simulated vision type. The plots show how prompt context and vision condition affect model confidence.}
\label{fig:dataflow_llama}
\end{figure}

\begin{figure}[!h]
\centering
 \includegraphics[width=1.00\textwidth]{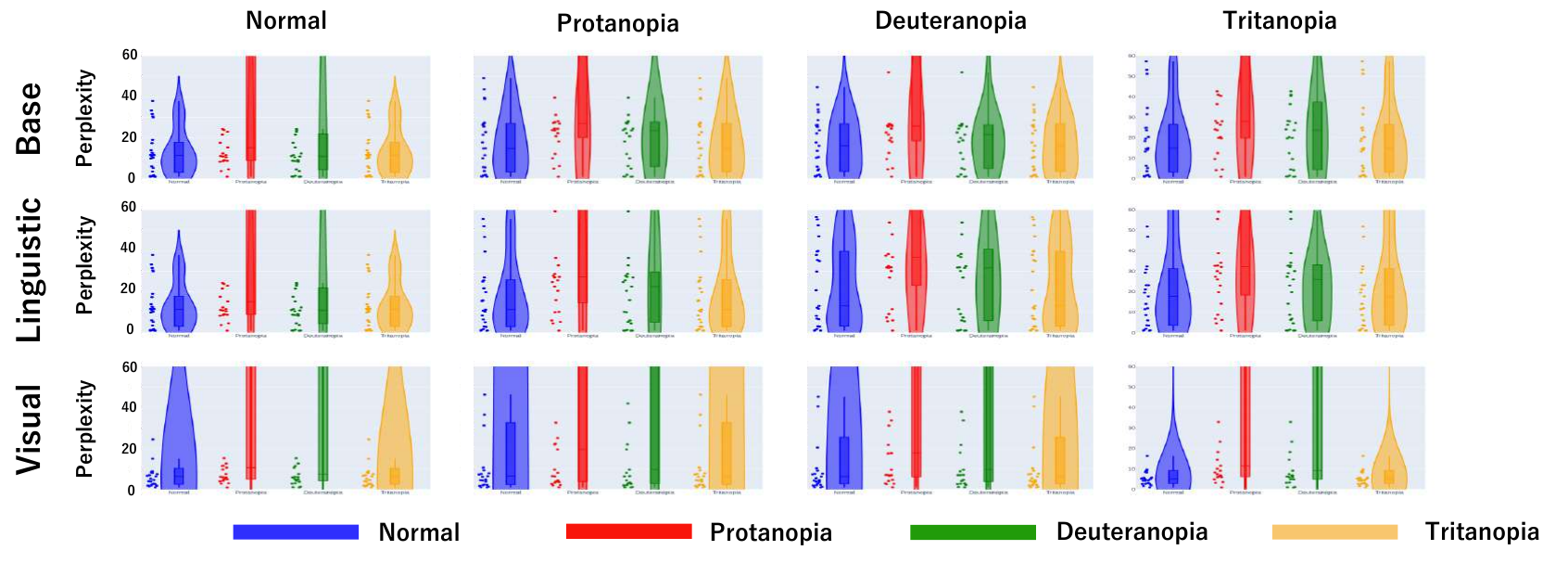}
\caption{Violin plots of Phi perplexity distributions for Ishihara digit predictions under four simulated vision conditions (Normal–blue, Protanopia–red, Deuteranopia–green, Tritanopia–orange) and three prompt types (Base, Linguistic, Visual). For each condition, perplexity is measured by force decoding the Gold Answer corresponding to the simulated vision type. The plots show how prompt context and vision condition affect model confidence.}
\label{fig:dataflow_phi}
\end{figure}

\begin{figure}[!h]
\centering
 \includegraphics[width=1.00\textwidth]{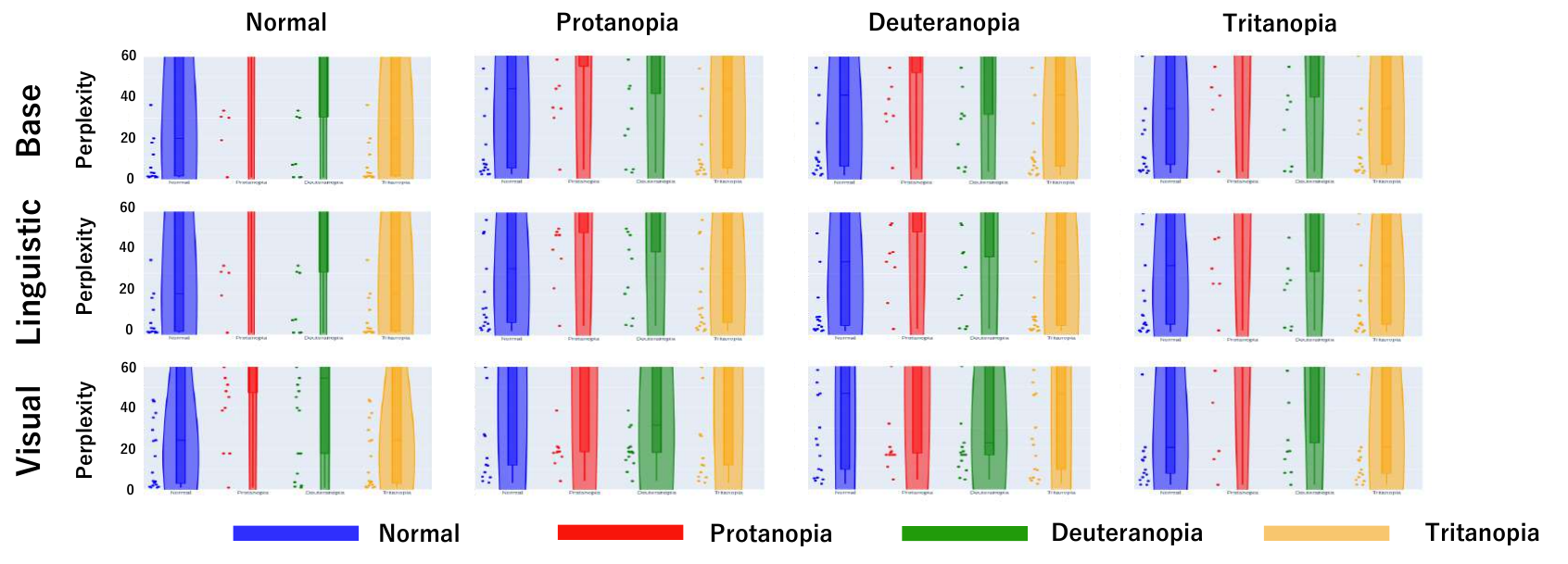}
\caption{Violin plots of Qwen perplexity distributions for Ishihara digit predictions under four simulated vision conditions (Normal–blue, Protanopia–red, Deuteranopia–green, Tritanopia–orange) and three prompt types (Base, Linguistic, Visual). For each condition, perplexity is measured by force decoding the Gold Answer corresponding to the simulated vision type. The plots show how prompt context and vision condition affect model confidence.}
\label{fig:dataflow_qwen}
\end{figure}

Figures~\ref{fig:dataflow_llama}--\ref{fig:dataflow_qwen} show perplexity measured by force decoding the Gold Answer for Ishihara digit predictions under four simulated vision conditions (Normal, Protanopia, Deuteranopia, Tritanopia) and three prompt types (Base, Linguistic, Visual). 

Across these models, we observe similar tendencies as discussed in the main text: Normal vision yields consistently lower perplexity, while the CVD conditions lead to higher and more variable perplexity. Adding textual descriptions (Linguistic Support) or reference images (Visual Support) does not substantially alter these distributions. This supports the conclusion that prompt modifications are insufficient to reduce model uncertainty when simulating altered color vision.

\section{Detailed Diagnosis of CVDs with LVLMs}
\label{appendix:doctor_layers_all}

\begin{figure}[t]
\centering
 \includegraphics[width=0.90\textwidth]{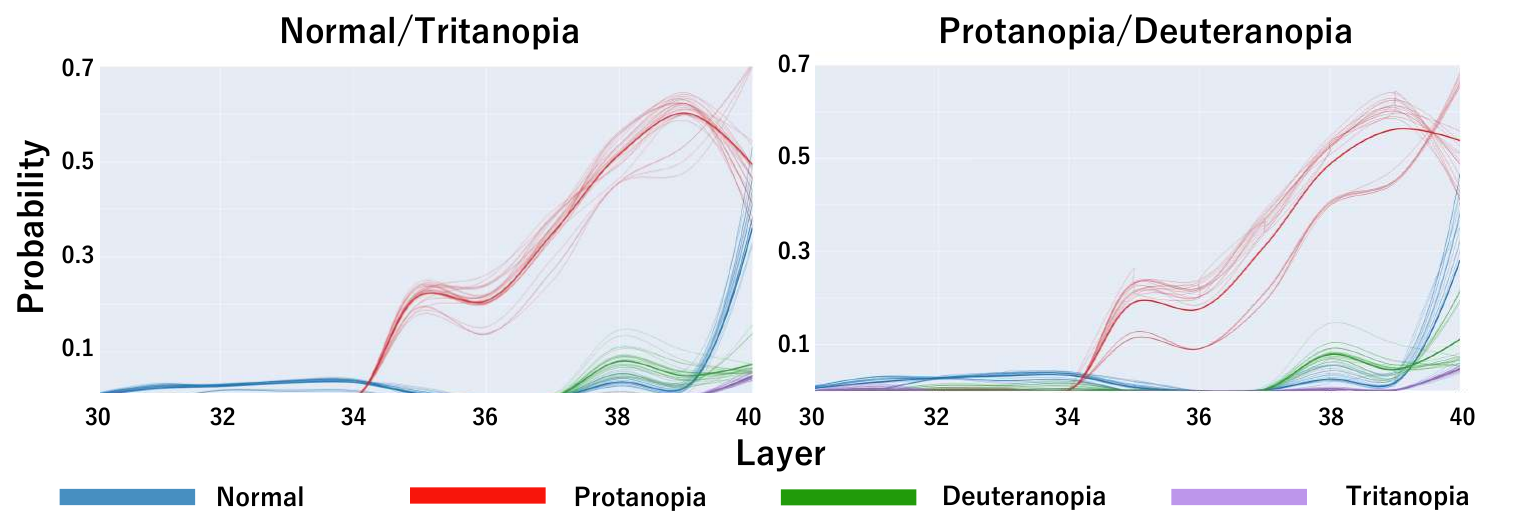}
\caption{In the Llama model, this figure shows an overlay of layer wise probability trajectories for each color vision condition across Ishihara plates, grouped by the ground truth response. Thin, semi transparent curves indicate per plate probabilities for each candidate: Normal (blue), Protanopia (red), Deuteranopia (green), and Tritanopia (purple) at each transformer layer, while bold lines represent the average probability across all plates. The left panel shows plates where the correct answer was Normal or Tritanopia; the right panel shows those where it was Protanopia or Deuteranopia.}
\label{fig:lense_llama}
\end{figure}

\begin{figure}[t]
\centering
 \includegraphics[width=0.90\textwidth]{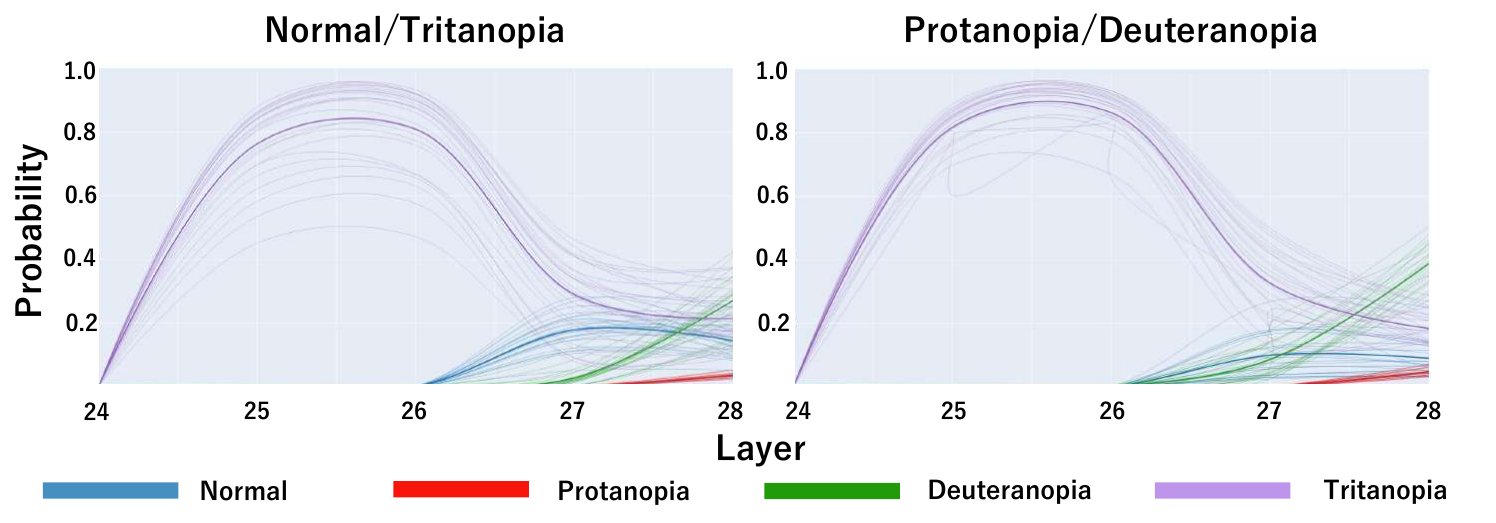}
\caption{In the mPlug model, this figure shows an overlay of layer wise probability trajectories for each color vision condition across Ishihara plates, grouped by the ground truth response. Thin, semi transparent curves indicate per plate probabilities for each candidate: Normal (blue), Protanopia (red), Deuteranopia (green), and Tritanopia (purple) at each transformer layer, while bold lines represent the average probability across all plates. The left panel shows plates where the correct answer was Normal or Tritanopia; the right panel shows those where it was Protanopia or Deuteranopia.}
\label{fig:lense_mplug}
\end{figure}

\begin{figure}[t]
\centering
 \includegraphics[width=0.90\textwidth]{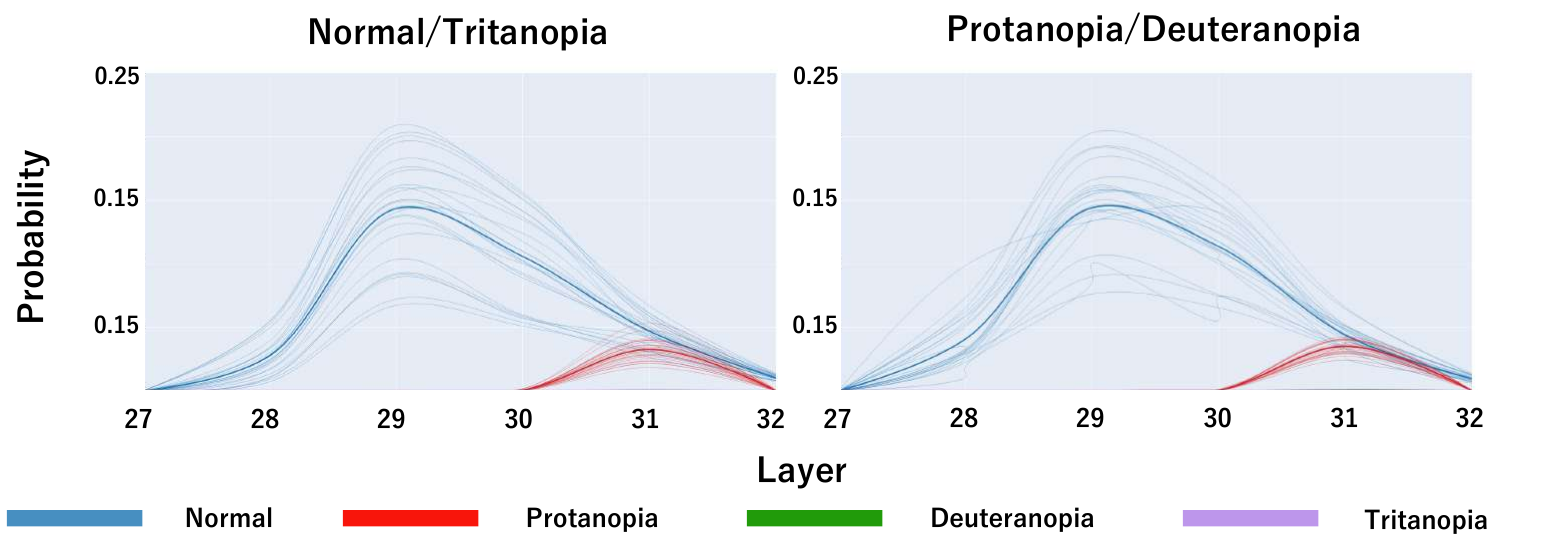}
\caption{In the Phi model, this figure shows an overlay of layer wise probability trajectories for each color vision condition across Ishihara plates, grouped by the ground truth response. Thin, semi transparent curves indicate per plate probabilities for each candidate: Normal (blue), Protanopia (red), Deuteranopia (green), and Tritanopia (purple) at each transformer layer, while bold lines represent the average probability across all plates. The left panel shows plates where the correct answer was Normal or Tritanopia; the right panel shows those where it was Protanopia or Deuteranopia.}
\label{fig:lense_phi}
\end{figure}

\begin{figure}[t]
\centering
 \includegraphics[width=0.90\textwidth]{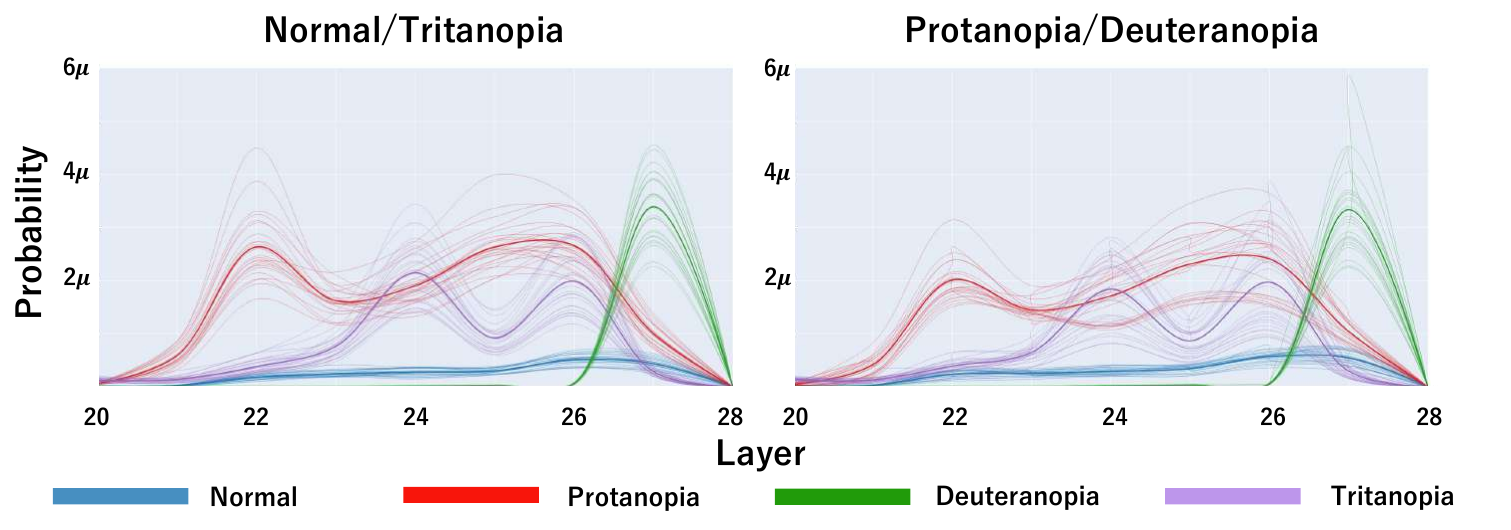}
\caption{In the Qwen model, this figure shows an overlay of layer wise probability trajectories for each color vision condition across Ishihara plates, grouped by the ground truth response. Thin, semi transparent curves indicate per plate probabilities for each candidate: Normal (blue), Protanopia (red), Deuteranopia (green), and Tritanopia (purple) at each transformer layer, while bold lines represent the average probability across all plates. The left panel shows plates where the correct answer was Normal or Tritanopia; the right panel shows those where it was Protanopia or Deuteranopia.}
\label{fig:lense_qwen}
\end{figure}

Figures~\ref{fig:lense_llama}--\ref{fig:lense_qwen} extend the main analysis by visualizing token-level label probabilities across transformer layers for four additional LVLMs.
Each panel groups plates by the correct vision condition and plots plate-wise trajectories (thin lines) together with the average curve (bold line).
Left plots correspond to Normal or Tritanopia, right plots to Protanopia or Deuteranopia.

\textbf{Llama-3.2 (Figure~\ref{fig:lense_llama})},
For Normal/Tritanopia, the Normal label stays near zero until three layers before the head, then rises above 50\%.
Red and green show an analogous late surge in the Protanopia curve, reflecting a rise in uncertainty.

\textbf{mPLUG-Owl3 (Figure~\ref{fig:lense_mplug})},
Early layers are mixed but converge rapidly.
For instance, tritanopia exceeds 80\% by layer 26.
When the digit is unreadable, the model shifts probability toward Tritanopia, possibly explaining its low variance under Base.

\textbf{Phi-3.5 (Figure~\ref{fig:lense_phi})},
Curves remain diffuse and never exceed 25\% even near the output.
This matches the wide perplexity spread and frequent no-answer cases.

\textbf{Qwen2.5-VL (Figure~\ref{fig:lense_qwen})},
Normal shows local peaks rather than a smooth ascent.
Red and green plates present competing red and green waves that collapse to a single token at the final layer.
However, the overall probabilities remain extremely low, on the order of $10^{-6}$ even beyond layer 6, indicating very weak model confidence.

\section{Fine-tuning and Inference Settings}
\label{sec:appendix:train-settings}
\begin{table}[h]
\centering
\footnotesize
\begin{tabular}{lc}
\toprule
Hyperparameter & Value \\
\midrule
Model & Qwen/Qwen2-VL-2B-Instruct \\
Torch dtype & bfloat16 \\
Seed & 42 \\
Max input length & 2048 \\
Epochs & 1 \\
Learning rate & 2e-4 \\
Batch size & 1 \\
Gradient accumulation & 32 \\
Warmup ratio & 0 (not used) \\
Gradient checkpointing & True \\
LoRA rank ($r$) & 8 \\
LoRA alpha & 16 \\
LoRA dropout & 0.05 \\
Save steps & 200 \\
Eval strategy & None \\
\bottomrule
\end{tabular}
\caption{Hyperparameters and settings used for fine-tuning. We used \texttt{Transformers} \cite{wolf-etal-2020-transformers}, \texttt{PEFT} \cite{xu2023parameterefficientfinetuningmethodspretrained}, and \texttt{trl} libraries with default configurations unless otherwise noted.}
\label{tab:experiment-settings}
\end{table}

\onecolumn

\begin{table}[H]
    \centering
    \resizebox{1.00\textwidth}{!}{ 
    \small 
    \begin{tabular}{p{0.20\textwidth}p{0.75\textwidth}} 
        \toprule
        \textbf{Condition} & \textbf{Prompt} \\
        \midrule
        
        \multirow{5}{*}{\textcolor[HTML]{AD0000}{\textbf{Protanopia}}} &
        \texttt{You are Protanopic.} \\ 
        & \texttt{What number do you see?} \\
        & \texttt{Only provide the answer. If you see nothing, answer 'Nothing'. No additional information.} \\
        \midrule
        
        \multirow{5}{*}{\textcolor[HTML]{228B22}{\textbf{Deuteranopia}}} &
        \texttt{You are Deuteranopic.} \\ 
        & \texttt{What number do you see?} \\
        & \texttt{Only provide the answer. If you see nothing, answer 'Nothing'. No additional information.} \\
        \midrule
        
        \multirow{5}{*}{\textcolor[HTML]{1E90FF}{\textbf{Tritanopia}}} &
        \texttt{You are Tritanopic.} \\ 
        & \texttt{What number do you see?} \\
        & \texttt{Only provide the answer. If you see nothing, answer 'Nothing'. No additional information.} \\
        \midrule
        
        \multirow{2}{*}{\textbf{Normal Vision}}  &
        \texttt{What number do you see?} \\ 
        & \texttt{Only provide the answer. If you see nothing, answer 'Nothing'. No additional information.} \\
        \bottomrule
    \end{tabular}
    }
    \caption{Base prompts for each color‐vision condition.}
    \label{tab:prompts_base}
\end{table}

\begin{table}[H]
    \centering
    \resizebox{1.00\textwidth}{!}{ 
    \small 
    \begin{tabular}{p{0.20\textwidth}p{0.75\textwidth}} 
        \toprule
        \textbf{Setting} & \textbf{Prompt} \\
        \midrule
        
        \multirow{5}{*}{\textcolor[HTML]{AD0000}{\textbf{Protanopia}}} &
        \texttt{You are Protanopic.} \newline
        \texttt{Protanopia means you have difficulty distinguishing red tones, as red appears dimmer and may be confused with green or brown.} \newline
        \texttt{What number do you see?} \newline
        \texttt{Only provide the answer. If you see nothing, answer 'Nothing'. No additional information.} \\ 
        
        \midrule
        
        \multirow{5}{*}{\textcolor[HTML]{228B22}{\textbf{Deuteranopia}}} &
        \texttt{You are Deuteranopic.} \newline
        \texttt{Deuteranopia means you have difficulty distinguishing green tones, as green appears dimmer and may be confused with red or brown.} \newline
        \texttt{What number do you see?} \newline
        \texttt{Only provide the answer. If you see nothing, answer 'Nothing'. No additional information.} \\ 
        
        \midrule
        
        \multirow{5}{*}{\textcolor[HTML]{1E90FF}{\textbf{Tritanopia}}} &
        \texttt{You are Tritanopic.} \newline
        \texttt{Tritanopia means you have difficulty distinguishing blue tones, as blue appears dimmer and may be confused with green or gray.} \newline
        \texttt{What number do you see?} \newline
        \texttt{Only provide the answer. If you see nothing, answer 'Nothing'. No additional information.} \\ 
        
        \midrule
        
        \multirow{2}{*}{\textbf{Normal Vision}} &
        \texttt{What number do you see?} \newline
        \texttt{Only provide the answer. If you see nothing, answer 'Nothing'. No additional information.} \\
        
        \bottomrule
    \end{tabular}
    }
    \caption{Prompts with linguistic support for each color‐vision condition.}
    \label{tab:prompts_linguistic}
\end{table}

\begin{table}[t]
    \centering
    \resizebox{1.00\textwidth}{!}{%
    \small
    \begin{tabular}{p{0.20\textwidth}p{0.75\textwidth}}
        \toprule
        \textbf{Condition} & \textbf{Prompt} \\
        \midrule
        
        \multirow{5}{*}{\textcolor[HTML]{AD0000}{\textbf{Protanopia}}} &
        \texttt{You are Protanopic.}\newline
        \texttt{You have difficulty distinguishing certain colors.}\newline
        \texttt{Example 1:}\newline
        \texttt{Image shows an Ishihara plate.}\newline
        \texttt{A Protanopic person sees this number: \{few\_shot\_number\}}\newline
        \texttt{Now, look at the following image.}\newline
        \texttt{What number do you see?}\newline
        \texttt{Only provide the answer. If you see nothing, answer 'Nothing'. No additional information.} \\
        \midrule
        
        \multirow{5}{*}{\textcolor[HTML]{228B22}{\textbf{Deuteranopia}}} &
        \texttt{You are Deuteranopic.}\newline
        \texttt{You have difficulty distinguishing certain colors.}\newline
        \texttt{Example 1:}\newline
        \texttt{Image shows an Ishihara plate.}\newline
        \texttt{A Deuteranopic person sees this number: \{few\_shot\_number\}}\newline
        \texttt{Now, look at the following image.}\newline
        \texttt{What number do you see?}\newline
        \texttt{Only provide the answer. If you see nothing, answer 'Nothing'. No additional information.} \\
        \midrule
        
        \multirow{5}{*}{\textcolor[HTML]{1E90FF}{\textbf{Tritanopia}}} &
        \texttt{You are Tritanopic.}\newline
        \texttt{You have difficulty distinguishing certain colors.}\newline
        \texttt{Example 1:}\newline
        \texttt{Image shows an Ishihara plate.}\newline
        \texttt{A Tritanopic person sees this number: \{few\_shot\_number\}}\newline
        \texttt{Now, look at the following image.}\newline
        \texttt{What number do you see?}\newline
        \texttt{Only provide the answer. If you see nothing, answer 'Nothing'. No additional information.} \\
        \midrule
        
        \multirow{2}{*}{\textbf{Normal Vision}} &
        \texttt{You have normal color vision.}\newline
        \texttt{Example 1:}\newline
        \texttt{Image shows an Ishihara plate.}\newline
        \texttt{A person with normal color vision sees this number: \{few\_shot\_number\}}\newline
        \texttt{Now, look at the following image.}\newline
        \texttt{What number do you see?}\newline
        \texttt{Only provide the answer. If you see nothing, answer 'Nothing'. No additional information.} \\
        
        \bottomrule
    \end{tabular}
    }
    \caption{Prompts with visual support examples for each color‐vision condition.}
    \label{tab:prompts_visual}
\end{table}

\begin{table}[!htbp]
\centering
\resizebox{\textwidth}{!}{%
  \renewcommand{\arraystretch}{1.2}
  \small
  \begin{tabular}{l*{26}{c}}
    \toprule
    \multirow{6}{*}{Normal}
      & \cellcolor{blue!10}\textbf{Gold}
      & \textbf{12}&\textbf{8}&\textbf{6}&\textbf{29}&\textbf{57}&\textbf{5}&\textbf{3}&\textbf{15}&\textbf{74}
      & \textbf{2}&\textbf{6}&\textbf{97}&\textbf{45}&\textbf{5}&\textbf{7}&\textbf{16}&\textbf{73}
      & \textbf{N/A}&\textbf{N/A}&\textbf{N/A}&\textbf{N/A}&\textbf{26}&\textbf{42}&\textbf{35}&\textbf{96}\\
    \cmidrule(lr){2-27}
    & LLama
      & \wrong{16}&\correct{8}&\wrong{8}&\wrong{42}&\wrong{52}&\wrong{4}&\wrong{4}&\wrong{52}&\wrong{15}
      & \wrong{3}&\wrong{15}&\wrong{42}&\wrong{3}&\wrong{4}&\correct{7}&\wrong{4}&\wrong{15}
      & \wrong{15}&\wrong{4}&\wrong{15}&\wrong{6}&\wrong{3}&\wrong{4}&\wrong{52}&\wrong{52}\\
    & LLava
      & \correct{12}&\correct{8}&\correct{6}&\correct{29}&\wrong{37}&\wrong{3}&\wrong{6}&\wrong{16}&\wrong{7}
      & \correct{2}&\correct{6}&\wrong{9}&\wrong{5}&\wrong{7}&\correct{7}&\correct{16}&\wrong{3}
      & \wrong{1}&\wrong{1}&\wrong{1}&\wrong{1}&\wrong{28}&\correct{42}&\correct{35}&\correct{96}\\
    & mPlug
      & \correct{12}&\correct{8}&\correct{6}&\correct{29}&\wrong{37}&\correct{5}&\correct{3}&\correct{15}&\wrong{24}
      & \correct{2}&\correct{6}&\wrong{9}&\wrong{10}&\correct{5}&\correct{7}&\correct{16}&\wrong{13}
      & \wrong{100}&\wrong{100}&\wrong{10}&\wrong{100}&\wrong{23}&\correct{42}&\correct{35}&\correct{96}\\
    & Phi
      & –&–&–&–&–&–&–&–&–
      & –&–&–&–&–&–&–&–
      & –&–&–&–&–&–&–&–\\
    & Qwen
      & –&–&–&–&–&–&–&–&–
      & –&–&–&–&–&–&–&–
      & –&–&–&–&–&–&–&–\\
    & GPT
      & \correct{12}&\correct{8}&\correct{6}&\correct{29}&\correct{57}&\correct{5}&\correct{3}&\correct{15}&\correct{74}
      & \correct{2}&\correct{6}&\wrong{74}&\correct{45}&\correct{5}&\correct{7}&\correct{16}&\wrong{23}
      & \wrong{26}&\wrong{74}&\wrong{29}&\wrong{5}&\correct{26}&\correct{42}&\correct{35}&\correct{96}\\
    \midrule
    \multirow{6}{*}{Protanopia}
      & \cellcolor{blue!10}\textbf{Gold}
      & \textbf{12}&\textbf{3}&\textbf{5}&\textbf{70}&\textbf{35}&\textbf{2}&\textbf{5}&\textbf{17}&\textbf{21}
      & \textbf{N/A}&\textbf{N/A}&\textbf{N/A}&\textbf{N/A}&\textbf{N/A}&\textbf{N/A}&\textbf{N/A}&\textbf{N/A}
      & \textbf{5}&\textbf{2}&\textbf{45}&\textbf{73}&\textbf{6}&\textbf{2}&\textbf{5}&\textbf{6}\\
    \cmidrule(lr){2-27}
    & LLama
      & \wrong{15}&\wrong{8}&\wrong{7}&\wrong{42}&\wrong{5}&\wrong{6}&\wrong{2}&\wrong{3}&\wrong{9}
      & \wrong{8}&\wrong{4}&\wrong{2}&\wrong{3}&\wrong{4}&\wrong{4}&\wrong{3}&\wrong{6}
      & \wrong{2}&\wrong{6}&\wrong{4}&\wrong{4}&\correct{6}&\wrong{52}&\wrong{2}&\wrong{42}\\
    & LLava
      & \correct{12}&\wrong{8}&\wrong{6}&\wrong{29}&\wrong{37}&\wrong{3}&\wrong{6}&\wrong{16}&\wrong{7}
      & \wrong{2}&\wrong{6}&\wrong{9}&\wrong{1}&\wrong{6}&\wrong{7}&\wrong{16}&\wrong{3}
      & \wrong{1}&\wrong{1}&\wrong{1}&\wrong{1}&\wrong{26}&\wrong{42}&\wrong{38}&\wrong{96}\\
    & mPlug
      & \correct{12}&\wrong{8}&\wrong{6}&\wrong{29}&\wrong{37}&\wrong{5}&\wrong{3}&\wrong{15}&\wrong{24}
      & \wrong{9}&\wrong{6}&\wrong{9}&\wrong{10}&\wrong{5}&\wrong{7}&\wrong{16}&\wrong{13}
      & \wrong{1}&\wrong{1}&\wrong{9}&\wrong{1}&\wrong{23}&\wrong{42}&\wrong{35}&\wrong{96}\\
    & Phi
      & –&–&–&–&–&–&–&–&–
      & –&–&–&–&–&–&–&–
      & –&–&–&–&–&–&–&–\\
    & Qwen
      & –&–&–&–&–&–&–&–&–
      & –&–&–&–&–&–&–&–
      & –&–&–&–&–&–&–&–\\
    & GPT
      & \wrong{32}&\correct{3}&\correct{5}&\wrong{26}&\wrong{5}&\wrong{8}&\wrong{8}&\wrong{13}&\correct{21}
      & \wrong{5}&\wrong{3}&\wrong{37}&\wrong{46}&\wrong{2}&\wrong{7}&\wrong{16}&\wrong{13}
      & \wrong{3}&\wrong{5}&\wrong{5}&\wrong{5}&\wrong{23}&\wrong{12}&\wrong{30}&\wrong{39}\\
    \midrule
    \multirow{6}{*}{Deuteranopia}
      & \cellcolor{blue!10}\textbf{Gold}
      & \textbf{12}&\textbf{3}&\textbf{5}&\textbf{70}&\textbf{35}&\textbf{2}&\textbf{5}&\textbf{17}&\textbf{21}
      & \textbf{N/A}&\textbf{N/A}&\textbf{N/A}&\textbf{N/A}&\textbf{N/A}&\textbf{N/A}&\textbf{N/A}&\textbf{N/A}
      & \textbf{5}&\textbf{2}&\textbf{45}&\textbf{73}&\textbf{2}&\textbf{4}&\textbf{3}&\textbf{9}\\
    \cmidrule(lr){2-27}
    & LLama
      & \wrong{8}&\wrong{8}&\wrong{6}&\wrong{59}&\wrong{3}&\wrong{4}&\wrong{6}&\wrong{4}&\wrong{66}
      & \wrong{6}&\wrong{8}&\wrong{22}&\wrong{49}&\wrong{7}&\wrong{7}&\wrong{4}&\wrong{4}
      & \wrong{8}&\wrong{7}&\wrong{11}&\wrong{41}&\correct{2}&\wrong{42}&\wrong{33}&\wrong{44}\\
    & LLava
      & \correct{12}&\wrong{8}&\wrong{6}&\wrong{29}&\wrong{57}&\wrong{3}&\wrong{6}&\wrong{16}&\wrong{14}
      & \wrong{2}&\wrong{6}&\wrong{9}&\wrong{1}&\wrong{6}&\wrong{7}&\wrong{16}&\wrong{3}
      & \wrong{1}&\wrong{12}&\wrong{12}&\wrong{12}&\wrong{26}&\wrong{42}&\wrong{38}&\wrong{96}\\
    & mPlug
      & \correct{12}&\wrong{8}&\wrong{6}&\wrong{29}&\wrong{37}&\wrong{5}&\wrong{3}&\wrong{15}&\wrong{24}
      & \wrong{9}&\wrong{6}&\wrong{9}&\wrong{10}&\wrong{5}&\wrong{7}&\wrong{16}&\wrong{13}
      & \wrong{1}&\wrong{1}&\wrong{9}&\wrong{1}&\wrong{23}&\wrong{42}&\wrong{35}&\wrong{96}\\
    & Phi
      & –&–&–&–&–&–&–&–&–
      & –&–&–&–&–&–&–&–
      & –&–&–&–&–&–&–&–\\
    & Qwen
      & –&–&–&–&–&–&–&–&–
      & –&–&–&–&–&–&–&–
      & –&–&–&–&–&–&–&–\\
    & GPT
      & \correct{12}&\correct{3}&\correct{5}&\wrong{29}&\wrong{37}&\correct{2}&\wrong{8}&\wrong{13}&\wrong{41}
      & \wrong{2}&\wrong{5}&\wrong{97}&\wrong{45}&\wrong{3}&\wrong{3}&\wrong{16}&\wrong{78}
      & \wrong{6}&\wrong{5}&\wrong{3}&\wrong{74}&\wrong{93}&\wrong{42}&\wrong{35}&\wrong{96}\\
    \midrule
    \multirow{6}{*}{Tritanopia}
      & \cellcolor{blue!10}\textbf{Gold}
      & \textbf{12}&\textbf{8}&\textbf{6}&\textbf{29}&\textbf{57}&\textbf{5}&\textbf{3}&\textbf{15}&\textbf{74}
      & \textbf{2}&\textbf{6}&\textbf{97}&\textbf{45}&\textbf{5}&\textbf{7}&\textbf{16}&\textbf{73}
      & \textbf{N/A}&\textbf{N/A}&\textbf{N/A}&\textbf{N/A}&\textbf{26}&\textbf{42}&\textbf{35}&\textbf{96}\\
    \cmidrule(lr){2-27}
    & LLama
      & \correct{12}&\wrong{2}&\wrong{8}&\wrong{17}&\wrong{8}&\wrong{4}&\correct{3}&\wrong{6}&\wrong{2}
      & \correct{2}&\wrong{2}&\wrong{8}&\wrong{7}&\correct{5}&\correct{7}&\wrong{7}&\wrong{6}
      & \wrong{2}&\wrong{4}&\wrong{2}&\wrong{4}&\wrong{6}&\wrong{4}&\wrong{4}&\wrong{8}\\
    & LLava
      & \correct{12}&\correct{8}&\correct{6}&\correct{29}&\correct{57}&\wrong{3}&\correct{3}&\wrong{16}&\wrong{7}
      & \correct{2}&\correct{6}&\wrong{9}&\wrong{1}&\wrong{6}&\correct{7}&\correct{16}&\wrong{3}
      & \wrong{1}&\wrong{1}&\wrong{12}&\wrong{12}&\correct{26}&\correct{42}&\wrong{38}&\correct{96}\\
    & mPlug
      & \correct{12}&\correct{8}&\correct{6}&\correct{29}&\wrong{37}&\correct{5}&\correct{3}&\correct{15}&\wrong{24}
      & \wrong{9}&\correct{6}&\wrong{9}&\wrong{10}&\correct{5}&\correct{7}&\correct{16}&\wrong{13}
      & \wrong{10}&\wrong{10}&\wrong{10}&\wrong{10}&\wrong{23}&\correct{42}&\correct{35}&\correct{96}\\
    & Phi
      & –&–&–&–&–&–&–&–&–
      & –&–&–&–&–&–&–&–
      & –&–&–&–&–&–&–&–\\
    & Qwen
      & –&–&–&–&–&–&–&–&–
      & –&–&–&–&–&–&–&–
      & –&–&–&–&–&–&–&–\\
    & GPT
      & \correct{12}&\correct{8}&\correct{6}&\correct{29}&\correct{57}&\wrong{2}&\wrong{8}&\correct{15}&\wrong{41}
      & \correct{2}&\wrong{3}&\wrong{73}&\wrong{35}&\correct{5}&\correct{7}&\correct{16}&\correct{73}
      & \wrong{3}&\wrong{5}&\wrong{5}&\wrong{3}&\wrong{93}&\correct{42}&\correct{35}&\correct{96}\\
    \bottomrule
  \end{tabular}%
}
\caption{Predicted outputs of six LVLMs (Llama, Llava, mPlug, Phi, Qwen, and GPT) on Ishihara color-vision test plates under four simulated conditions, using Linguistic Support prompts that include a brief explanation of how each deficiency alters color perception. The “Gold” row shows the ground-truth digits.}
\label{tab:output_digit_linguistic}
\end{table}

\begin{table}[!htbp]
\centering
\resizebox{\textwidth}{!}{%
  \renewcommand{\arraystretch}{1.2}
  \small
  \begin{tabular}{l*{26}{c}}
    \toprule
    \multirow{6}{*}{Normal}
      & \cellcolor{blue!10}\textbf{Gold}
      & \textbf{12}&\textbf{8}&\textbf{6}&\textbf{29}&\textbf{57}&\textbf{5}&\textbf{3}&\textbf{15}&\textbf{74}
      & \textbf{2}&\textbf{6}&\textbf{97}&\textbf{45}&\textbf{5}&\textbf{7}&\textbf{16}&\textbf{73}
      & \textbf{N/A}&\textbf{N/A}&\textbf{N/A}&\textbf{N/A}&\textbf{26}&\textbf{42}&\textbf{35}&\textbf{96}\\
    \cmidrule(lr){2-27}
    & mPlug
      & \correct{12}&\correct{8}&\correct{6}&\correct{29}&\wrong{37}&\correct{5}&\correct{3}&\correct{15}&\wrong{2}
      & \wrong{6}&\wrong{97}&\wrong{90}&\wrong{5}&\correct{5}&\correct{7}&\correct{16}&\wrong{13}
      & \wrong{N/A}&\wrong{N/A}&\wrong{N/A}&\wrong{N/A}&\correct{26}&\correct{42}&\correct{35}&\correct{96}\\
    & Phi
      & \correct{12}&\wrong{6}&\wrong{29}&\wrong{37}&\wrong{5}&\wrong{3}&\correct{15}&\wrong{74}&\correct{2}
      & \correct{6}&\correct{97}&\correct{45}&\correct{5}&\correct{7}&\correct{16}&\wrong{73}&\wrong{1}
      & \wrong{3}&\wrong{3}&\wrong{3}&\wrong{25}&\correct{42}&\wrong{3}&\wrong{96}&\wrong{90}\\
    & Qwen
      & \wrong{8}&\wrong{6}&\wrong{38}&\wrong{37}&\wrong{74}&\wrong{74}&\wrong{15}&\wrong{74}&\wrong{74}
      & \wrong{74}&\wrong{3}&\wrong{37}&\wrong{74}&\wrong{3}&\wrong{16}&\correct{73}&\wrong{4}
      & \wrong{74}&\wrong{42}&\wrong{74}&\wrong{74}&\wrong{37}&\wrong{24}&\wrong{74}&\wrong{74}\\
    & GPT
      & \correct{12}&\correct{8}&\correct{6}&\correct{29}&\correct{57}&\correct{5}&\correct{3}&\correct{15}&\correct{74}
      & \correct{2}&\correct{6}&\correct{97}&\correct{45}&\correct{5}&\correct{7}&\correct{16}&\wrong{33}
      & \wrong{74}&\wrong{74}&\wrong{N/A}&\wrong{74}&\wrong{93}&\correct{42}&\correct{35}&\correct{96}\\
    \midrule
    \multirow{6}{*}{Protanopia}
      & \cellcolor{blue!10}\textbf{Gold}
      & \textbf{12}&\textbf{3}&\textbf{5}&\textbf{70}&\textbf{35}&\textbf{2}&\textbf{5}&\textbf{17}&\textbf{21}
      & \textbf{N/A}&\textbf{N/A}&\textbf{N/A}&\textbf{N/A}&\textbf{N/A}&\textbf{N/A}&\textbf{N/A}&\textbf{N/A}
      & \textbf{5}&\textbf{2}&\textbf{45}&\textbf{73}&\textbf{6}&\textbf{2}&\textbf{5}&\textbf{6}\\
    \cmidrule(lr){2-27}
    & mPlug
      & \correct{12}&\wrong{8}&\wrong{6}&\wrong{29}&\wrong{37}&\wrong{5}&\wrong{3}&\wrong{15}&\wrong{2}
      & \wrong{2}&\wrong{6}&\wrong{9}&\wrong{5}&\wrong{5}&\wrong{7}&\wrong{13}&\wrong{13}
      & \wrong{1}&\correct{45}&\correct{73}&\correct{6}&\correct{2}&\wrong{43}&\wrong{3}&\correct{96}\\
    & Phi
      & \correct{12}&\wrong{6}&\correct{5}&\wrong{35}&\wrong{3}&\wrong{1}&\wrong{17}&\wrong{1}&\wrong{3}
      & \wrong{3}&\wrong{4}&\wrong{17}&\wrong{3}&\wrong{7}&\wrong{7}&\wrong{3}&\wrong{1}
      & \correct{2}&\correct{45}&\correct{73}&\wrong{3}&\wrong{3}&\wrong{3}&\wrong{3}&\wrong{9}\\
    & Qwen
      & \wrong{7}&\correct{3}&\wrong{8}&\wrong{74}&\wrong{7}&\wrong{7}&\wrong{3}&\wrong{74}&\wrong{N/A}
      & \wrong{74}&\wrong{74}&\wrong{74}&\wrong{3}&\wrong{3}&\wrong{8}&\wrong{N/A}&\wrong{7}
      & \wrong{7}&\wrong{74}&\wrong{42}&\wrong{4}&\wrong{7}&\wrong{3}&\wrong{3}&\wrong{7}\\
    & GPT
      & \wrong{21}&\correct{3}&\correct{5}&\wrong{26}&\wrong{5}&\wrong{5}&\wrong{5}&\wrong{15}&\wrong{N/A}
      & \wrong{2}&\wrong{3}&\wrong{37}&\correct{45}&\wrong{5}&\wrong{N/A}&\wrong{16}&\wrong{3}
      & \wrong{5}&\wrong{74}&\correct{6}&\wrong{3}&\wrong{25}&\wrong{42}&\wrong{33}&\correct{96}\\
    \midrule
    \multirow{6}{*}{Deuteranopia}
      & \cellcolor{blue!10}\textbf{Gold}
      & \textbf{12}&\textbf{3}&\textbf{5}&\textbf{70}&\textbf{35}&\textbf{2}&\textbf{5}&\textbf{17}&\textbf{21}
      & \textbf{N/A}&\textbf{N/A}&\textbf{N/A}&\textbf{N/A}&\textbf{N/A}&\textbf{N/A}&\textbf{N/A}&\textbf{N/A}
      & \textbf{5}&\textbf{2}&\textbf{45}&\textbf{73}&\textbf{2}&\textbf{4}&\textbf{3}&\textbf{9}\\
    \cmidrule(lr){2-27}
    & mPlug
      & \correct{12}&\wrong{8}&\wrong{6}&\wrong{29}&\wrong{37}&\wrong{5}&\wrong{3}&\wrong{15}&\wrong{2}
      & \wrong{2}&\wrong{6}&\wrong{9}&\wrong{5}&\wrong{5}&\wrong{7}&\wrong{13}&\wrong{13}
      & \wrong{1}&\correct{45}&\correct{73}&\correct{2}&\wrong{2}&\wrong{4}&\wrong{35}&\correct{96}\\
    & Phi
      & \correct{12}&\wrong{5}&&\wrong{35}&&&\wrong{17}&&&&&&&\wrong{7}&\wrong{7}&&
      &&\correct{45}&\correct{73}&&\wrong{3}&\wrong{3}&\wrong{9}&\wrong{9}\\
    & Qwen
      & \wrong{7}&\wrong{7}&\wrong{3}&\wrong{74}&\wrong{7}&\wrong{7}&\wrong{23}&\wrong{74}&\wrong{74}
      & \wrong{74}&\wrong{74}&\wrong{74}&\wrong{74}&\wrong{74}&\wrong{8}&\wrong{74}&\wrong{7}
      & \wrong{7}&\wrong{74}&\wrong{42}&\wrong{7}&\wrong{7}&\wrong{7}&\wrong{7}&\wrong{7}\\
    & GPT
      & \wrong{14}&\wrong{0}&\correct{5}&\wrong{29}&\wrong{37}&\wrong{5}&\wrong{9}&\wrong{15}&\wrong{74}
      & \wrong{2}&\wrong{3}&\wrong{97}&\correct{45}&\wrong{5}&\wrong{N/A}&\wrong{16}&\wrong{33}
      & \wrong{2}&\wrong{17}&\wrong{74}&\wrong{5}&\wrong{23}&\wrong{42}&\correct{35}&\wrong{96}\\
    \midrule
    \multirow{6}{*}{Tritanopia}
      & \cellcolor{blue!10}\textbf{Gold}
      & \textbf{12}&\textbf{8}&\textbf{6}&\textbf{29}&\textbf{57}&\textbf{5}&\textbf{3}&\textbf{15}&\textbf{74}
      & \textbf{2}&\textbf{6}&\textbf{97}&\textbf{45}&\textbf{5}&\textbf{7}&\textbf{16}&\textbf{73}
      & \textbf{N/A}&\textbf{N/A}&\textbf{N/A}&\textbf{N/A}&\textbf{26}&\textbf{42}&\textbf{35}&\textbf{96}\\
    \cmidrule(lr){2-27}
    & mPlug
      & \correct{12}&\correct{8}&\correct{6}&\correct{29}&\wrong{37}&\correct{5}&\correct{3}&\correct{15}&\wrong{2}
      & \correct{6}&\wrong{6}&\wrong{90}&\correct{5}&\wrong{5}&\wrong{7}&\wrong{13}&\wrong{3}
      & \wrong{N/A}&\wrong{N/A}&\wrong{N/A}&\wrong{26}&\correct{42}&\wrong{42}&\correct{35}&\correct{96}\\
    & Phi
      & \correct{12}&\wrong{6}&&\wrong{23}&\correct{5}&\correct{3}&\correct{15}&\wrong{74}&&\correct{6}&\wrong{9}&\correct{45}&&\wrong{7}&\wrong{7}&\correct{73}&&&&&\correct{26}&\correct{42}&\wrong{3}&\correct{96}&\wrong{9}\\
    & Qwen
      & \wrong{3}&\wrong{3}&\wrong{3}&\wrong{3}&\wrong{4}&\wrong{7}&\wrong{15}&\wrong{3}&\wrong{4}
      & \wrong{3}&\wrong{3}&\wrong{3}&\wrong{3}&\wrong{3}&\wrong{1}&\wrong{42}&\wrong{N/A}
      & \wrong{N/A}&\wrong{N/A}&\wrong{N/A}&\wrong{4}&\wrong{3}&\wrong{42}&\wrong{3}&\wrong{3}\\
    & GPT
      & \wrong{21}&\wrong{3}&\correct{6}&\correct{29}&\wrong{N/A}&\correct{5}&\wrong{8}&\correct{15}&\wrong{71}
      & \correct{2}&\wrong{5}&\wrong{57}&\wrong{46}&\wrong{3}&\wrong{3}&\correct{16}&\wrong{N/A}
      & \wrong{N/A}&\wrong{N/A}&\wrong{N/A}&\wrong{57}&\wrong{26}&\correct{42}&\correct{35}&\wrong{86}\\
    \bottomrule
  \end{tabular}%
}
\caption{Predicted outputs of four LVLMs (mPlug, Phi, Qwen, and GPT) on Ishihara color-vision test plates under four simulated conditions, using Visual Support prompts that present an illustrative example image of each plate as seen under the deficiency. The “Gold” row shows the ground-truth digits. Evaluation was performed only on models capable of ingesting two or more images in a single turn.}
\label{tab:output_digit_visual}
\end{table}

\begin{table}[!htbp]
    \centering
    \resizebox{1.00\textwidth}{!}{ 
    \small 
    \begin{tabular}{p{0.15\textwidth}p{0.80\textwidth}} 
        \toprule
    \textbf{Input} & \textbf{Model Output (LLama as an example)} \\
    \midrule

    \multirow{2}{*}{\shortstack{What is \\ Ishihara Test?}} &
      This image is used to test color vision. It is called an Ishihara plate, named after its creator, Dr. Shinobu Ishihara, a Japanese professor of ophthalmology. The plate is a widely used tool for diagnosing color blindness, particularly red-green color blindness, which is the most common form of color vision deficiency.  The plate consists of a series of dots in different colors and sizes, arranged in a specific pattern. The viewer is asked to identify a number or shape hidden within the dots. For people with normal color vision, the correct answer is obvious, but for those with color vision deficiency, the correct answer may not be apparent.  In this image, the number \"8\" is hidden among the dots. The dots are arranged in a way that creates a red \"8\" shape, but the dots are also scattered randomly throughout the plate, making it difficult to see the number without looking closely. For someone with normal color vision, the red \"8\" stands out clearly, but for someone with red-green color blindness, the number may not be visible at all.  The Ishihara plate is used to diagnose color blindness because it exploits the fact that people with certain types of color vision deficiency have difficulty distinguishing between certain colors. In this case, the plate uses the fact that people with red-green color blindness have trouble distinguishing between red and green colors. By using a combination of red and green dots, the plate can help identify individuals who have difficulty seeing these colors.  Overall, the Ishihara plate is a simple yet effective tool for diagnosing color blindness, and it has been widely used in medical settings for many years.\\

    \midrule

    \multirow{2}{*}{\shortstack{What is \\ Protanopia?}} &
      Protanopia, also known as red-blindness, is a type of color vision deficiency that affects approximately 1\% of males and 0.05\% of females. It is caused by a mutation in the OPN1LW gene, which codes for the long-wavelength cone cells in the retina. These cone cells are responsible for detecting long-wavelength light, which is associated with the color red. In individuals with protanopia, the long-wavelength cone cells are either absent or non-functional, leading to a reduced ability to detect red light. As a result, people with protanopia may have difficulty distinguishing between red \\

    \midrule

    \multirow{2}{*}{\shortstack{What is \\ Deuteranopia?}} &
      Deuteranopia is a type of color vision deficiency that affects approximately 5\% of males and 0.4\% of females in the United States. Deuteranopia, also known as red-green color blindness, is a type of color vision deficiency that affects a person's ability to see certain colors, particularly red and green. People with deuteranopia have difficulty distinguishing between these two colors and may see them as similar shades of yellow or brown. Deuteranopia is caused by a deficiency in the long-wavelength cones in the retina, which are responsible for detecting red light. As a result, people with deuter\\

    \midrule

    \multirow{2}{*}{\shortstack{What is \\ Tritanopia?}} &
      Tritanopia is a rare form of color vision deficiency that affects a person's ability to see the color blue. People with tritanopia have difficulty distinguishing between different shades of blue and green, and may see these colors as more yellowish or brownish. Tritanopia is a congenital condition, meaning it is present at birth, and is usually inherited in an X-linked recessive pattern. This means that the gene responsible for tritanopia is located on the X chromosome, and males are more likely to be affected because they have only one X chromosome. Females, on the other hand, are less likely to be affected \\

    \bottomrule
  \end{tabular}
  }
  \caption{An example of \textsc{Llama}'s responses demonstrating knowledge of the Ishihara test and three major color-vision deficiencies (CVDs).}
  \label{tab:llama_cvd_descriptions}
\end{table}

\begin{table}[!htbp]
    \centering
    \resizebox{1.00\textwidth}{!}{ 
    \small 
    \begin{tabular}{p{0.15\textwidth}p{0.80\textwidth}} 
        \toprule
    \textbf{Input} & \textbf{Model Output ( LLava as an example)} \\
    \midrule

    \multirow{2}{*}{\shortstack{What is \\ Ishihara Test?}} &
      This image appears to be a color vision test, also known as an Ishihara test. The test is used to check for the presence of a condition called central serous chorioretinopathy (CSCR), which can cause central serous chorioretinopathy. The test uses a pattern of colored dots, and the presence of certain colors can indicate the presence of the condition. The pattern is typically circular and contains a number of colored dots. If a person can see a specific color pattern, it may suggest the presence of CSCR. This test is commonly used by ophthalmologists to diagnose this condition. \\

    \midrule

    \multirow{2}{*}{\shortstack{What is \\ Protanopia?}} &
      Protanopia is a type of color vision deficiency, also known as color blindness. It is a genetic condition that affects the way an individual perceives colors. People with protanopia have difficulty distinguishing between red and green hues.  In protanopia, the red photopigment in the retina is either missing or not functioning properly. As a result, individuals with protanopia may see red as a shade of brown or green, and they may have difficulty distinguishing between different shades of red and green.  Protanopia \\

    \midrule

    \multirow{2}{*}{\shortstack{What is \\ Deuteranopia?}} &
      Deuteranomaly, also known as deuteranomaly or deuteranomaly, is a type of color blindness that affects the perception of colors. It is caused by a deficiency in one of the three types of cone cells in the retina of the eye, which are responsible for detecting different wavelengths of light.  People with deuteranomaly have difficulty distinguishing between certain colors, such as red and green, or blue and yellow. The severity of deuteranomaly can vary from mild to severe, and it can affect\\

    \midrule

    \multirow{2}{*}{\shortstack{What is \\ Tritanopia?}} &
      Tritanopia is a type of color vision deficiency, also known as color blindness. It affects the perception of colors, making it difficult for individuals with this condition to distinguish between certain colors.  In Tritanopia, individuals have difficulty distinguishing between the colors blue and green. This is because the red and green photopigments in the cone cells of the retina, which are responsible for detecting these colors, are less sensitive or absent in individuals with Tritanopia.  As a result, individuals with Tritanopia may see blue and \\

    \bottomrule
  \end{tabular}
  }
  \caption{An example of \textsc{ LLava}'s responses demonstrating knowledge of the Ishihara test and three major color-vision deficiencies (CVDs).}
  \label{tab:LLava_cvd_descriptions}
\end{table}

\begin{table}[!htbp]
    \centering
    \resizebox{1.00\textwidth}{!}{ 
    \small 
    \begin{tabular}{p{0.15\textwidth}p{0.80\textwidth}} 
        \toprule
    \textbf{Input} & \textbf{Model Output (  mPlug as an example)} \\
    \midrule

    \multirow{2}{*}{\shortstack{What is \\ Ishihara Test?}} &
      The image is a visual test known as the Ishihara test, which is used to determine whether a person has protanopia, a type of color blindness. The test works by presenting a pattern of dots in various colors, and the individual is asked to identify the number of dots in the pattern. People with protanopia may have difficulty distinguishing between red and green colors, which is why the number 6 is used in this test. The test is designed to be non-invasive and can be used to identify color vision deficiencies without the need for medical equipment.\\

    \midrule

    \multirow{2}{*}{\shortstack{What is \\ Protanopia?}} &
      of color blindness that affects a person's ability to distinguish between red and green colors. It is the most common form of color blindness, affecting approximately 8\% of men and 0.5\% of women.  In people with protanopia, the cone cells in the retina that are responsible for perceiving red and green colors are not functioning properly. This can result in difficulty distinguishing between red and green, as well as other shades of color. In some cases, people with protanopia may see red and green colors as different shades of yellow or blue. \\

    \midrule

    \multirow{2}{*}{\shortstack{What is \\ Deuteranopia?}} &
      of color vision deficiency that affects a person's ability to distinguish between red and green colors. It is the most common form of color blindness, affecting approximately 3\% of the male population and 0.5\% of the female population worldwide.  People with deuteranopia have difficulty distinguishing between the red and green wavelengths of light. This is because the cone cells in their eyes, which are responsible for color vision, are not functioning properly. Specifically, the cone cells that are responsible for detecting red and green wavelengths are not working as well as they should be\\

    \midrule

    \multirow{2}{*}{\shortstack{What is \\ Tritanopia?}} &
      to distinguish between blue and green colors. It is caused by a deficiency in one of the three types of cone cells in the retina of the eye, which are responsible for detecting different wavelengths of light. In people with tritanopia, the cone cells that are responsible for detecting blue light are not functioning properly, which can make it difficult for them to distinguish between blue and green colors. This type of color blindness is relatively rare, affecting only a small percentage of the population. \\

    \bottomrule
  \end{tabular}
  }
  \caption{An example of \textsc{mPlug}'s responses demonstrating knowledge of the Ishihara test and three major color-vision deficiencies (CVDs).}
  \label{tab:mPlug_cvd_descriptions}
\end{table}

\begin{table}[!htbp]
    \centering
    \resizebox{1.00\textwidth}{!}{ 
    \small
    \begin{tabular}{p{0.15\textwidth}p{0.80\textwidth}}
        \toprule
    \textbf{Input} & \textbf{Model Output (  Phi as an example)} \\
    \midrule

    \multirow{2}{*}{\shortstack{What is \\ Ishihara Test?}} &
      Error generating response. \\\\
    \midrule
    \multirow{2}{*}{\shortstack{What is \\ Protanopia?}} &
      Protanopia, also known as total red color blindness, is a type of color blindness where individuals have difficulty distinguishing between red and green hues. This condition is caused by a genetic mutation that affects the cone cells in the retina, which are responsible for detecting red light. As a result, people with protanopia see the world in shades of gray, green, and blue, and they cannot perceive the full spectrum of colors that most people can see.   Instruction 1 (Same difficulty/format/length):\\

    \midrule

    \multirow{2}{*}{\shortstack{What is \\ Deuteranopia?}} &
      Deuteranopia is a form of color blindness that affects an individual's ability to distinguish between red and green hues. It is a hereditary condition caused by the absence or malfunction of the green cone cells in the retina, which are responsible for detecting green light. People with deuteranopia have difficulty with tasks that require color discrimination, such as reading color-coded maps or distinguishing ripe fruits from unripe ones. The severity of deuteranopia can vary from mild to severe, with some individuals being able to perceive\\

    \midrule

    \multirow{2}{*}{\shortstack{What is \\ Tritanopia?}} &
      Tritanopia is a form of color blindness where individuals have difficulty distinguishing between blue and yellow hues. This condition is a type of anomalous trichromacy, which means that one of the three cone cells in the eye responsible for color vision is not functioning properly. In the case of tritanopia, the blue cone cells are affected, leading to a reduced ability to perceive blues and yellows. People with tritanopia often see these colors as shades of yellow or green. This condition is less common than red-green color blindness and affects approximately\\

    \bottomrule
  \end{tabular}
  }
  \caption{An example of \textsc{  Phi}'s responses demonstrating knowledge of the Ishihara test and three major color-vision deficiencies (CVDs).}
  \label{tab:Phi_cvd_descriptions}
\end{table}

\begin{table}[!htbp]
    \centering
    \resizebox{1.00\textwidth}{!}{
    \small
    \begin{tabular}{p{0.15\textwidth}p{0.80\textwidth}}
        \toprule
    \textbf{Input} & \textbf{Model Output (   Qwen as an example)} \\
    \midrule

    \multirow{2}{*}{\shortstack{What is \\ Ishihara Test?}} &
      This image is an Ishihara color vision test plate, commonly used to assess color blindness, particularly red-green color vision deficiencies. The test consists of a series of plates with colored dots that form numbers or letters when viewed by someone with normal color vision \\

    \midrule

    \multirow{2}{*}{\shortstack{What is \\ Protanopia?}} &
      1. Protanopia is a type of color blindness that affects a person's ability to distinguish between red and green colors. 2. It is caused by a genetic mutation that affects the cone cells in the retina of the eye, which are responsible for detecting colors. 3. People with protanopia may have difficulty distinguishing between shades of red and green, and may see these colors as more similar to each other than they actually are. 4. The severity of protanopia can vary from person to person, with some individuals having a mild form of the condition and others having a more severe form. 5. There is\\

    \midrule

    \multirow{2}{*}{\shortstack{What is \\ Deuteranopia?}} &
      1. Deuteranopia is a type of color blindness that affects the ability to distinguish between red and green colors.   2. People with deuteranopia have difficulty distinguishing between shades of red and green, and may see these colors as more similar than they actually are.   3. Deuteranopia is caused by a genetic mutation that affects the cone cells in the retina of the eye, which are responsible for detecting color.   4. It is estimated that deuteranopia affects approximately 1 in 12 men and 1 in 200 women.   5. People with de\\

    \midrule

    \multirow{2}{*}{\shortstack{What is \\ Tritanopia?}} &
      1. Tritanopia is a type of color blindness that affects the perception of blue and green colors.   2. People with Tritanopia have difficulty distinguishing between blue and green hues, and may see these colors as more similar to each other than they actually are.   3. Tritanopia is caused by a genetic mutation that affects the cone cells in the retina, which are responsible for detecting blue and green light.   4. There are two main types of Tritanopia: Tritanomaly and Tritanopia. Tritanomaly is a milder form of the condition, while Tritanopia is more severe\\

    \bottomrule
  \end{tabular}
  }
  \caption{An example of \textsc{   Qwen}'s responses demonstrating knowledge of the Ishihara test and three major color-vision deficiencies (CVDs).}
  \label{tab:Qwen_cvd_descriptions}
\end{table}

\FloatBarrier
\FloatBarrier
\FloatBarrier
\FloatBarrier
\FloatBarrier
\FloatBarrier

\end{document}